\pdfoutput=1
\documentclass[11pt,twoside,letterpaper]{article} %% The same for {book}
\usepackage{times,fancyhdr}
\usepackage[dvips]{graphicx}
\usepackage{algorithm2e}
\usepackage{amssymb}
\usepackage{amsmath}
\usepackage{bm}
\usepackage{url}
\usepackage{cite}
\usepackage{makeidx}
\makeindex
\usepackage{epstopdf}
\usepackage{color}
\usepackage{hyperref}
\newcommand{\vpi}{\mathbf{\pi}}
\newcommand{\rar}{\rightarrow}

\newcommand{\ttt}{\texttt}
\newcommand{\mbb}{\mathbb}
\newcommand{\mbf}{\mathbf}
\newcommand{\mca}{\mathcal}


\makeatletter 
\def\cleardoublepage{\clearpage\if@twoside \ifodd\c@page\else% 
    \hbox{}% 
    \thispagestyle{empty}%
    \newpage% 
    \if@twocolumn\hbox{}\newpage\fi\fi\fi} 
\makeatother

\begin{document}
\title{
{\begin{flushleft}
\vskip 0.45in
{\normalsize\bfseries}%\textit{Chapter~1}}
\end{flushleft}
\vskip 0.45in
\bfseries\scshape Interpretations of the Web of Data}}
\author{\bfseries\itshape Marko A. Rodriguez\thanks{E-mail address: marko@lanl.gov}\\
T-5, Center for Nonlinear Studies \\
Los Alamos National Laboratory \\
Los Alamos, New Mexico 87545}
\date{\today}

\maketitle

\begin{abstract}
The emerging Web of Data utilizes the web infrastructure to represent and interrelate data. The foundational standards of the Web of Data include the Uniform Resource Identifier (URI) and the Resource Description Framework (RDF). URIs are used to identify resources and RDF is used to relate resources. While RDF has been posited as a logic language designed specifically for knowledge representation and reasoning, it is more generally useful if it can conveniently support other models of computing. In order to realize the Web of Data as a general-purpose medium for storing and processing the world's data, it is necessary to separate RDF from its logic language legacy and frame it simply as a data model. Moreover, there is significant advantage in seeing the Semantic Web as a particular interpretation of the Web of Data that is focused specifically on knowledge representation and reasoning. By doing so, other interpretations of the Web of Data are exposed that realize RDF in different capacities and in support of different computing models.
\end{abstract}

\thispagestyle{empty}
\setcounter{page}{1}
% ------- [First Page Running Head] - place it immediately after title! ------
\thispagestyle{fancy}
\fancyhead{}
\fancyhead[L]{In: Data Management in the Semantic Web \\ 
Editor: Editor Name, pp. {\thepage-\pageref{lastpage-01}}} % needs \label{lastpage-01} on the last page.
\fancyhead[R]{ISBN 0000000000  \\
\copyright~2007 Nova Science Publishers, Inc.}
\fancyfoot{}
\renewcommand{\headrulewidth}{0pt}
%------------------------------------------------------------------------------

\vspace{2in}

% ------------ [Running Heads - for odd and even pages] - please insert it only on page 2!
\pagestyle{fancy}
\fancyhead{}
\fancyhead[EC]{Marko A. Rodriguez}
\fancyhead[EL,OR]{\thepage}
\fancyhead[OC]{Interpretations of the Web of Data}
\fancyfoot{}
\renewcommand\headrulewidth{0.5pt} 
%------------------------------------------------------------------------------

\section{Introduction}

The common conception of the World Wide Web is that of a large-scale, distributed file repository \cite{lee94}. The typical files found on the World Wide Web are Hyper-Text Markup Language (HTML) documents and other media such as image, video, and audio files. The ``World Wide'' aspect of the World Wide Web pertains to the fact that all of these files have an accessible location that is denoted by a Uniform Resource Locator (URL) \cite{uri:2001}; a URL denotes what physical machine is hosting the file (i.e.~what domain name/IP address), where in that physical machine the file is located (i.e.~what directory), and finally, which protocol to use to retrieve that file from that machine (e.g.~\ttt{http}, \ttt{ftp}, etc.). The ``Web'' aspect of the World Wide Web pertains to the fact that a file (typically an HTML document) can make reference (typically an \ttt{href} citation) to another file. In this way, a file on machine $A$ can link to a file on machine $B$ and in doing so, a network/graph/web of files emerges. The ingenuity of the World Wide Web is that it combines remote file access protocols and hypermedia and as such, has fostered a revolution in the way in which information is disseminated and retrieved---in an open, distributed manner. From this relatively simple foundation, a rich variety of uses emerges: from the homepage, to the blog, to the online store.

The World Wide Web is primarily for human consumption. While HTML documents are structured according to a machine understandable syntax, the content of the documents are written in human readable/writable language (i.e.~natural human language). It is only through computationally expensive and relatively inaccurate text analysis algorithms that a machine can determine the meaning of such documents. For this reason, computationally inexpensive keyword extraction and keyword-based search engines are the most prevalent means by which the World Wide Web is machine processed. However, the human-readable World Wide Web is evolving to support a machine-readable Web of Data. The emerging Web of Data utilizes the same referencing paradigm as the World Wide Web, but instead of being focused primarily on URLs and files, it is focused on Uniform Resource Identifiers (URI) \cite{uri:berners2005} and data.\footnote{The URI is the parent class of both the URL and the Uniform Resource Name (URN) \cite{uri:2001}.} The ``Data'' aspect of the Web of Data pertains to the fact that a URI can denote anything that can be assigned an identifier:  a physical entity, a virtual entity, an abstract concept, etc. The ``Web'' aspect of the Web of Data pertains to the fact that identified resource can be related to other resources by means of the Resource Description Framework (RDF). Among other things, RDF is an abstract data model that specifies the syntactic rules by which resources are connected. If $U$ is the set of all URIs, $B$ the set of all blank or anonymous nodes, and $L$ the set of all literals, then the Web of Data is defined as
%%%
\begin{equation*}
	\mca{W} \subseteq ((U \cup B) \times U \times (U \cup B \cup L)).
\end{equation*}
%%%
A single statement (or triple) in $\mca{W}$ is denoted $(s,p,o)$, where $s$ is called the subject, $p$ the predicate, and $o$ the object. On the Web of Data
%%%
\begin{quote}
``[any man or machine can] start with one data source and then move through a potentially endless Web of data sources connected by RDF links. Just as the traditional document Web can be crawled by following hypertext links, the Web of Data can be crawled by following RDF links. Working on the crawled data, search engines can provide sophisticated query capabilities, similar to those provided by conventional relational databases. Because the query results themselves are structured data, not just links to HTML pages, they can be immediately processed, thus enabling a new class of applications based on the Web of Data.'' \cite{linkeddata:bizer2008}
\end{quote}

As a data model, RDF can conveniently represent commonly used data structures. From the knowledge representation and reasoning perspective, RDF provides the means to make assertions about the world and infer new statements given existing statements. From the network/graph analysis perspective, RDF supports the representation of various network data structures. From the programming and systems engineering perspective, RDF can be used to encode objects, instructions, stacks, etc. The Web of Data, with its general-purpose data model and supporting technological infrastructure, provides various computing models a shared, global, distributed space. Unfortunately, this general-purpose, multi-model vision was not the original intention of the designers of RDF. RDF was created for the domain of knowledge representation and reasoning. Moreover, it caters to a particular monotonic subset of this domain \cite{rdfsem:hayes2004}. RDF is not generally understood as supporting different computing models. However, if the Web of Data is to be used as just that, a ``web of data'', then it is up to the applications leveraging this data to interpret what that data means and what it can be used for. 

The URI address space is an address space. It is analogous, in many ways, to the address space that exists in the local memory of the physical machines that support the representation of the Web of Data. With physical memory, information is contained at an address. For a 64-bit machine, that information is a 64-bit word. That 64-bit word can be interpreted as a literal primitive (e.g.~a byte, an integer, a floating point value) or yet another 64-bit address (i.e.~a pointer). This is how address locations denote data and link to each other, respectively. On the Web of Data, a URI is simply an address as it does not contain content.\footnote{This is not completely true. Given that a URL is a subtype of a URI, and a URL can ``contain'' a file, it is possible for a URI to ``contain'' information.} It is through RDF that a URI address has content. For instance, with RDF, a URI can reference a literal (i.e.~\ttt{xsd:byte}, \ttt{xsd:integer}, \ttt{xsd:float}) or another URI. Thus, RDF, as a data model, has many similarities to typical local memory. However, the benefit of URIs and RDF is that they create an inherently distributed and theoretically infinite space. Thus, the Web of Data can be interpreted as a large-scale, distributed memory structure. What is encoded and processed in that memory structure should not be dictated at the level of RDF, but instead dictated by the domains that leverage this medium for various application scenarios. The Web of Data should be realized as an application agnostic memory structure that supports a rich variety of uses: from Semantic Web reasoning, to Giant Global Graph analysis, to Web of Process execution.

The intention of this article is to create a conceptual splinter that separates RDF from its legacy use as a logic language and demonstrate that it is more generally applicable when realized as only a data model. In this way, RDF as the foundational standard for the Web of Data makes the Web of Data useful to anyone wishing to represent information and compute in a global, distributed space. Three specific interpretations of the Web of Data are presented in order to elucidate the many ways in which the Web of Data is currently being used. Moreover, within these different presentations, various standards and technologies are discussed. These presentations are provided as summaries, not full descriptions. In short, this article is more of a survey of a very large and multi-domained landscape. The three interpretations that will be discussed are enumerated below. 
%%%
\begin{enumerate}\addtolength{\itemsep}{-0.5\baselineskip}
	\item	The Web of Data as a knowledge base (see \S \ref{sec:knowledge-base}).
		\begin{itemize}
			\item The Semantic Web is an interpretation of the Web of Data.
			\item RDF is the means by which a model of a world is created.
			\item There are many types of logic: logics of truth and logics of thought.
			\item Scalable solutions exist for reasoning on the Web of Data.
		\end{itemize}
	\item The Web of Data as a multi-relational network (see \S \ref{sec:multi-relational}).
		\begin{itemize}
			\item The Giant Global Graph is an interpretation of the Web of Data.
			\item RDF is the means by which vertices are connected together by labeled edges.
			\item Single-relational network analysis algorithms can be applied to multi-relational networks.
			\item Scalable solutions exist for network analysis on the Web of Data.
		\end{itemize}
	\item The Web of Data as an object repository (see \S \ref{sec:object-repository}). 
		\begin{itemize}
			\item The Web of Process is an interpretation of the Web of Data.
			\item RDF is the means by which objects are represented and related to other objects.
			\item An object's representation can include both its fields and its methods.
			\item Scalable solutions exist for object-oriented computing on the Web of Data.
		\end{itemize}
\end{enumerate}
%%%
The landscape presented in this article is by no means complete and only provides a glimpse into these different areas. Moreover, within each of these three presented interpretations, applications and use-cases are not provided. What is provided is a presentation of common computing models that have been mapped to the Web of Data in order to take unique advantage of the Web as a computing infrastructure.

\section{A Distributed Knowledge Base}\label{sec:knowledge-base}

The Web of Data can be interpreted as a distributed knowledge base---a Semantic Web. A knowledge base is composed of a set of statements about some ``world''. These statements are written in some language. Inference rules designed for that language can be used to derive new statements from existing statements. In other words, inference rules can be used to make explicit what is implicit. This process is called reasoning. The Semantic Web initiative is primarily concerned with this interpretation of the Web of Data.
%%%
\begin{quote}
``For the Semantic Web to function, computers must have access to structured collections of information and sets of inference rules that they can use to conduct automated reasoning.'' \cite{lee:semantic2001}
\end{quote}
%%%
Currently, the Semantic Web interpretation of the Web of Data forces strict semantics on RDF. That is, RDF is not simply a data model, but a logic language. As a data model, it specifies how a statement $\tau$ is constructed (i.e.~$\tau \in ((U \cup B) \times U \times (U \cup B \cup L))$). As a logic language is species specific language constructs and semantics---a way of interpreting what statements mean. Because RDF was developed in concert with requirements provided by the knowledge representation and reasoning domain, RDF and the Semantic Web have been very strongly aligned for many years. This is perhaps the largest conceptual stronghold that exists as various W3C documents make this point explicit.
%%%
\begin{quote}
``RDF is an assertional logic, in which each triple expresses a simple proposition. This imposes a fairly strict monotonic discipline on the language, so that it cannot express closed-world assumptions, local default preferences, and several other commonly used non-monotonic constructs.'' \cite{rdfsem:hayes2004}
\end{quote}
%%%
RDF is monotonic in that any asserted statement $\tau \in \mca{W}$ can not be made ``false'' by future assertions. In other words, the truth-value of a statement, once stated, does not change. RDF makes use of the open-world assumption in that if a statement is not asserted, this does not entail that it is ``false''. The open-world assumption is contrasted to the closed-world assumption found in many systems, where the lack of data is usually interpreted as that data being ``false''.  

From this semantic foundation, extended semantics for RDF have been defined. The two most prevalent language extensions are the RDF Schema (RDFS) \cite{rdfs:brickley2004} and the Web Ontology Language (OWL) \cite{owlspec:mcguinness2004}. It is perhaps this stack of standards that forms the most common conception of what the Semantic Web is. However, if the Semantic Web is to be just that, a ``semantic web'', then there should be a way to represent other languages with different semantics. If RDF is forced to be a monotonic, open-world language, then this immediately pigeonholes what can be represented on the Semantic Web. If RDF is interpreted strictly as a data model, devoid of semantics, then any other knowledge representation language can be represented in RDF and thus, contribute to the Semantic Web. This section will discuss three logic languages: RDFS, OWL, and the Non-Axiomatic Logic (NAL) \cite{inherit:wang1994}. RDFS and OWL are generally understood in the Semantic Web community as these are the primary logic languages used. However, NAL is a multi-valent, non-monotonic language that, if to be implemented in the Semantic Web, requires that RDF be interpreted as a data model, not as a logic language. Moreover, NAL is an attractive language for the Semantic Web because its reasoning process is inherently distributed, can handle conflicting inconsistent data, and was designed on the assumption of insufficient knowledge and computing resources.

\subsection{RDF Schema}

\begin{figure}[h!]
	\begin{center}
		\includegraphics[width=0.65\textwidth]{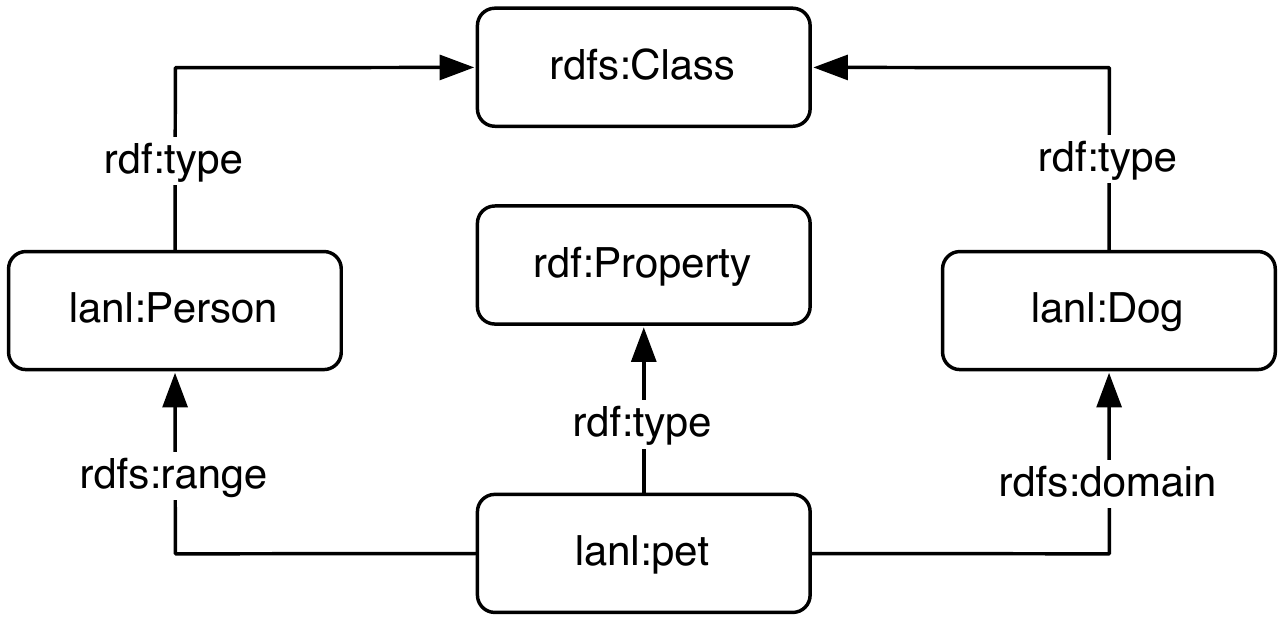}
	\caption{\label{fig:rdfs-example}An RDFS ontology that states that a person has a dog as a pet.}
	\end{center}
\end{figure}

RDFS is a simple language with a small set of inference rules \cite{rdfs:brickley2004}. In RDF, resources (e.g.~URIs and blank nodes) maintain properties (i.e.~\ttt{rdf:Property}). These properties are used to relate resources to other resources and literals. In RDFS, classes and properties can be formally defined. Class definitions organize resources into abstract categories. Property definitions specify the way in which these resources are related to one another. For example, it is possible to state there exist people and dogs (i.e.~classes) and people have dogs as pets (i.e.~a property). This is represented in RDFS in Figure \ref{fig:rdfs-example}.
%%%

RDFS inference rules are used to derive new statements given existing statements that use the RDFS langauge. RDFS inference rules make use of statements with the following URIs:
%%%
\begin{itemize}\addtolength{\itemsep}{-0.5\baselineskip}
	\item \ttt{rdfs:Class}: denotes a class as opposed to an instance.
	\item \ttt{rdf:Property}: denotes a property/role.
	\item \ttt{rdfs:domain}: denotes what a property projects from.
	\item \ttt{rdfs:range}: denotes what a property projects to.
	\item \ttt{rdf:type}: denotes that an instance is a type of class.
	\item \ttt{rdfs:subClassOf}: denotes that a class is a subclass of another.
	\item \ttt{rdfs:subPropertyOf}: denotes that a property is a sub-property of another.
	\item \ttt{rdfs:Resource}: denotes a generic resource.
	\item \ttt{rdfs:Datatype}: denotes a literal primitive class.
	\item \ttt{rdfs:Literal}: denotes a generic literal class.
\end{itemize}
%%%
RDFS supports two general types of inference: subsumption and realization. Subsumption determines which classes are a subclass of another. The RDFS inference rules that support subsumption are
%%%
\begin{equation*}
(?x, \ttt{rdf:type}, \ttt{rdfs:Class}) \implies (?x, \ttt{rdfs:subClassOf}, \ttt{rdfs:Resource}), 
\end{equation*}
%%%
\begin{equation*}
(?x, \ttt{rdf:type}, \ttt{rdfs:Datatype}) \implies (?x, \ttt{rdfs:subClassOf}, \ttt{rdfs:Literal}),
\end{equation*}
%%%
\begin{align*}
(?x, \ttt{rdfs:subPropertyOf}, ?y) \, \wedge \, &  (?y, \ttt{rdfs:subPropertyOf}, ?z) \\
& \implies (?x, \ttt{rdfs:subPropertyOf}, ?z). 
\end{align*}
%%%
and finally,
%%%
\begin{align*}
(?x, \ttt{rdfs:subClassOf}, ?y) \, \wedge \, &  (?y, \ttt{rdfs:subClassOf}, ?z) \\
& \implies (?x, \ttt{rdfs:subClassOf}, ?z). 
\end{align*}
%%%
Thus, if both

\begin{verbatim}
(lanl:Chihuahua, rdfs:subClassOf, lanl:Dog)
(lanl:Dog, rdfs:subClassOf, lanl:Mammal)
\end{verbatim}
%%%
are asserted, then it can be inferred that 

\begin{verbatim}
(lanl:Chihuahua, rdfs:subClassOf, lanl:Mammal).
\end{verbatim}
%%%
Next, realization is used to determine if a resource is an instance of a class. The RDFS inference rules that support realization are
%%%
\begin{equation*}
(?x, ?y, ?z) \implies (?x, \ttt{rdf:type}, \ttt{rdfs:Resource}), 
\end{equation*}
%%%
\begin{equation*}
(?x, ?y, ?z) \implies (?y, \ttt{rdf:type}, \ttt{rdf:Property}), 
\end{equation*}
%%%
\begin{equation*}
(?x, ?y, ?z) \implies (?z, \ttt{rdf:type}, \ttt{rdfs:Resource}), 
\end{equation*}
%%%
\begin{equation*}
(?x, \ttt{rdf:type}, ?y) \wedge(?y, \ttt{rdfs:subClassOf}, ?z)  \implies (?x, \ttt{rdf:type}, ?z),
\end{equation*}
%%%
\begin{equation*}
(?w, \ttt{rdfs:domain}, ?x) \wedge(?y, ?w, ?z) \implies (?y, \ttt{rdf:type}, ?x),
\end{equation*}
%%%
and finally,
%%%
\begin{equation*}
(?w, \ttt{rdfs:domain}, ?x) \wedge (?y, ?w, ?z) \implies (?z, \ttt{rdf:type}, ?x).
\end{equation*}
%%%
Thus if, along with the statements in Figure \ref{fig:rdfs-example},

\begin{verbatim}
(lanl:marko, lanl:pet, lanl:fluffy)
\end{verbatim}
%%%
is asserted, then it can be inferred that 

\begin{verbatim}
(lanl:marko, rdf:type, lanl:Person)
(lanl:fluffy, rdf:type, lanl:Dog).
\end{verbatim}

Given a knowledge base containing statements, these inference rules continue to execute until they no longer produce novel statements. It is the purpose of an RDFS reasoner to efficiently execute these rules. There are two primary ways in which inference rules are executed: at insert time and at query time. With respect to insert time, if a statement is inserted (i.e.~asserted) into the knowledge base, then the RDFS inference rules execute to determine what is entailed by this new statement. These newly entailed statements are then inserted in the knowledge base and the process continues. While this approach ensures fast query times (as all entailments are guaranteed to exist at query time), it greatly increases the number of statements generated. For instance, given a deep class hierarchy, if a resource is a type of one of the leaf classes, then it asserted that it is a type of all the super classes of that leaf class. In order to alleviate the issue of ``statement bloat,'' inference can instead occur at query time. When a query is executed, the reasoner determines what other implicit statements should be returned with the query. The benefits and drawbacks of each approach are benchmarked, like much of computing, according to space vs. time. 

\subsection{Web Ontology Language}

OWL is a more complicated language which extends RDFS by providing more expressive constructs for defining classes \cite{owlspec:mcguinness2004}. Moreover, beyond subsumption and realization, OWL provides inference rules to determine class and instance equivalence. There are many OWL specific inference rules. In order to give the flavor of OWL, without going into the many specifics, this subsection will only present some examples of the more commonly used constructs. For a fine, in depth review of OWL, please refer to \cite{owl:lacy2005}.

\begin{figure}[h!]
	\begin{center}
		\includegraphics[width=0.6\textwidth]{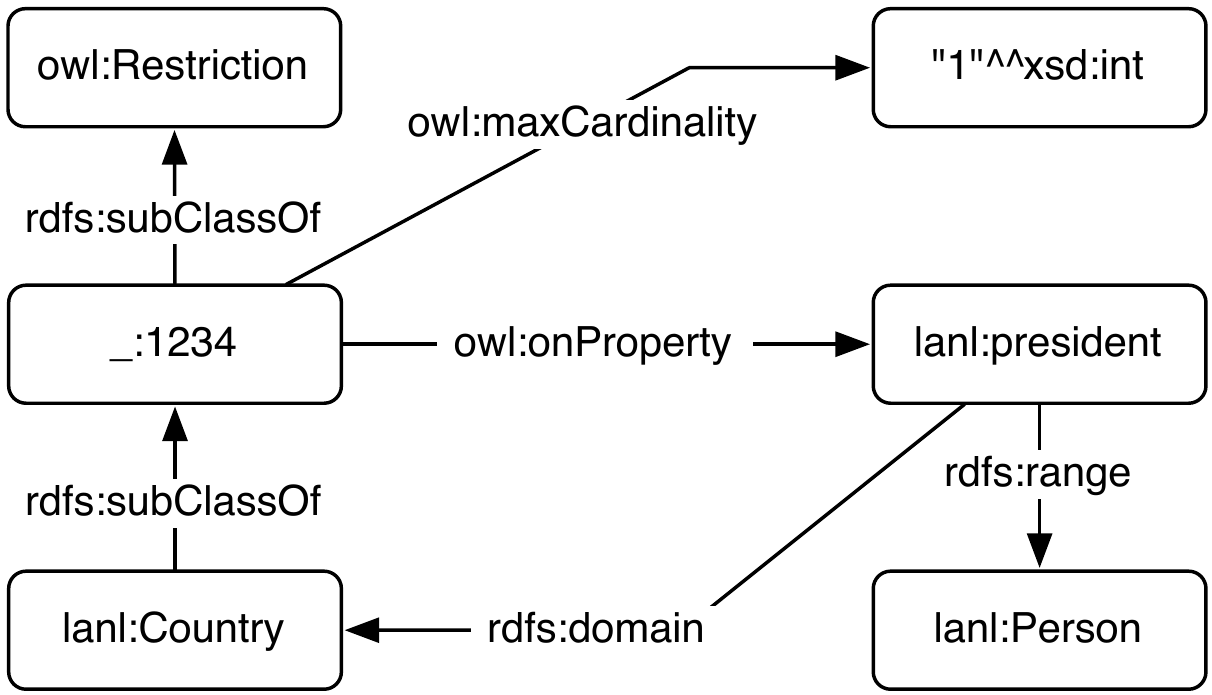}
	\caption{\label{fig:country-owl}An OWL ontology that states that the president of a country is a person and there can be at most one president for a country.}
	\end{center}
\end{figure}

Perhaps the most widely used language URI in OWL is \ttt{owl:Restriction}. In RDFS, a property can only have a domain and a range. In OWL, a class can apply the following restrictions to a property:
%%%
\begin{itemize}\addtolength{\itemsep}{-0.5\baselineskip}
	\item \ttt{owl:cardinality}
	\item \ttt{owl:minCardinality}
	\item \ttt{owl:maxCardinality}
	\item \ttt{owl:hasValue}
	\item \ttt{owl:allValuesFrom}
	\item \ttt{owl:someValuesFrom}
\end{itemize}
%%%
Cardinality restrictions are used to determine equivalence and inconsistency. For example, in an OWL ontology, it is possible to state that a country can only have one president. This is expressed in OWL as diagrammed in Figure \ref{fig:country-owl}. The \ttt{\_:1234} resource is a blank node that denotes a restriction on the country class's \ttt{lanl:president} property.
%%%

Next, if \ttt{usa:barack} and \ttt{usa:obama} are both asserted to be the president of the United States with the statements

\begin{verbatim}
(usa:barack, lanl:president, usa:United_States)
(usa:obama, lanl:president, usa:United_States),
\end{verbatim} 
%%%
then it can be inferred (according to OWL rules) that these resources are equivalent. This equivalence relationship is made possible because the maximum cardinality of the \ttt{lanl:president} property of a country is $1$. Therefore, if there are ``two'' people that are president, then they must be the same person. This is made explicit when the reasoner asserts the statements

\begin{verbatim}
(usa:barack, owl:sameAs, usa:obama)
(usa:obama, owl:sameAs, usa:barack).
\end{verbatim} 
%%%
Next, if \ttt{lanl:herbertv} is asserted to be different from \ttt{usa:barack} (which, from previous, was asserted to be the same as \ttt{usa:obama}) and \ttt{lanl:herbertv} is also asserted to be the president of the United States, then an inconsistency is detected. Thus, given the ontology asserted in Figure \ref{fig:country-owl} and the previous assertions, asserting

\begin{verbatim}
(lanl:herbertv, owl:differentFrom, usa:barack)
(lanl:herbertv, lanl:president, usa:United_States)
\end{verbatim} 
%%%
causes an inconsistency. This inconsistency is due to the fact that a country can only have one president and \ttt{lanl:herbertv} is not \ttt{usa:barack}. 

Two other useful language URIs for properties in OWL are 
%%%
\begin{itemize}\addtolength{\itemsep}{-0.5\baselineskip}
	\item \ttt{owl:SymmetricProperty}
	\item \ttt{owl:TransitiveProperty}
	
\end{itemize}
%%%
In short, if $y$ is symmetric, then if $(x,y,z)$ is asserted, then $(z,y,x)$ can be inferred. Next, if the property $y$ is transitive, then if $(w,y,x)$ and $(x,y,z)$ are asserted then, $(w,y,z)$ can be inferred. 

There are various reasoners that exist for the OWL language. A popular OWL reasoner is Pellet \cite{pellet:2004}. The purpose of Pellet is to execute the OWL rules given existing statements in the knowledge base. For many large-scale knowledge base applications (i.e.~triple- or quad-stores), the application provides its own reasoner. Popular knowledge bases that make use of the OWL language are OWLim \cite{owlim:kiryakov2005}, Oracle Spatial \cite{oracle:alexander2006}, and AllegroGraph \cite{agraph:aasman2006}. It is noted that due to the complexity (in terms of implementation and running times), many knowledge base reasoners only execute subsets of the OWL language. For instance, AllegroGraph's reasoner is called RDFS++ as it implements all of the RDFS rules and only some of the OWL rules. However, it is also noted that RacerPro \cite{racer:haarslev2003} can be used with AllegroGraph to accomplish complete OWL reasoning. Finally, OpenSesame \cite{sesame:dutchy2002} can be used for RDFS reasoning. Because OpenSesame is both a knowledge base and an API, knowledge base applications that implement the OpenSesame interfaces can automatically leverage the OpenSesame RDFS reasoner; though there may be speed issues as the reasoner is not natively designed for that knowledge base application. 

\subsection{Non-Axiomatic Logic}

If RDF is strictly considered a monotonic, open-world logic language, then the Semantic Web is solidified as an open-world, monotonic logic environment. If reasoning is restricted to the legacy semantics of RDF, then it will become more difficult to reason on the Semantic Web as it grows in size and as more inconsistent knowledge is introduced. With the number of statements of the Semantic Web, computational hurdles are met when reasoning with RDFS and OWL. With inconsistent statements on the Semantic Web, it is difficult to reason as inconsistencies are not handled gracefully in RDFS or OWL. In general, sound and complete reasoning will not be feasible as the Semantic Web continues to grow. In order to meet these challenges, the Large Knowledge Collider project (LarKC) is focused on developing a reasoning platform to handle incomplete and inconsistent data \cite{larkc:fensel2008}.
%%%
\begin{quote}
``Researchers have developed methods for reasoning in rather small, closed, trustworthy, consistent, and static domains. They usually provide a small set of axioms and facts. [OWL] reasoners can deal with $10^5$ axioms (concept definitions), but they scale poorly for large instance sets. [...] There is a deep mismatch between reasoning on a Web scale and efficient reasoning algorithms over restricted subsets of first-order logic. This is rooted in underlying assumptions of current systems for computational logic: small set of axioms, small number of facts, completeness of inference, correctness of inference rules and consistency, and static domains.'' \cite{larkc:fensel2008}
\end{quote}

There is a need for practical methods to reason on the Semantic Web. One promising logic was founded on the assumption of insufficient knowledge and resources. This logic is called the Non-Axiomatic Logic (NAL) \cite{nars2.2:wang}. Unfortunately for the Semantic Web as it is now, NAL breaks the assumptions of RDF semantics as NAL is multi-valent, non-monotonic, and makes use of statements with a subject-predicate form. However, if RDF is considered simply a data model, then it is possible to represent NAL statements and make use of its efficient, distributed reasoning system. Again, for the massive-scale, inconsistent world of the Semantic Web, sound and complete approaches are simply becoming more unreasonable. 

\subsubsection{Language}

There are currently 8 NAL languages. Each language, from NAL-0 to NAL-8, builds on the constructs of the previous in order to support more complex statements. The following list itemizes the various languages and what can be expressed in each.
%%%
\begin{itemize}\addtolength{\itemsep}{-0.5\baselineskip}
	\item NAL-0: binary inheritance.
	\item NAL-1: inference rules.
	\item NAL-2: sets and variants of inheritance.
	\item NAL-3: intersections and differences
	\item NAL-4: products, images, and ordinary relations.
	\item NAL-5: statement reification.
	\item NAL-6: variables.
	\item NAL-7: temporal statements.
	\item NAL-8: procedural statements.
\end{itemize}

Every NAL language is based on a simple inheritance relationship. For example, in NAL-0, which assumes all statements are binary,
%%%
\begin{equation*}
\ttt{lanl:marko} \rar \ttt{lanl:Person}
\end{equation*}
%%%
states that Marko (subject) inherits ($\rar$) from person (predicate). Given that all subjects and predicates are joined by inheritance, there is no need to represent the copula when formally representing a statement.\footnote{This is not completely true as different types of inheritance are defined in NAL-2 such as instance $\circ\!\!\!\!\rar$, property $\rar\!\!\!\circ$, and instance-property $\circ\!\!\!\!\rar\!\!\!\circ$ inheritance. However, these $3$ types of inheritance can also be represented using the basic $\rar$ inheritance. Moreover, the RDF representation presented can support the explicit representation of other inheritance relationships if desired.}. If RDF, as a data model, is to represent NAL, then one possible representation for the above statement is

\begin{verbatim}
(lanl:marko, lanl:1234, lanl:Person),
\end{verbatim}
%%%
where \ttt{lanl:1234} serves as a statement pointer. This pointer could be, for example, a 128-bit Universally Unique Identifier (UUID) \cite{uuid:leach2005}. It is important to maintain a statement pointer as beyond NAL-0, statements are not simply ``true'' or ``false''. A statement's truth is not defined by its existence, but instead by extra numeric metadata associated with the statement. NAL maintains an
%%%
\begin{quote}
``experience-grounded semantics [where] the truth value of a judgment indicates the degree to which the judgment is supported by the system's experience. Defined in this way, truth value is system-dependent and time-dependent. Different systems may have conflicting opinions, due to their different experiences.'' \cite{inherit:wang1994}
\end{quote}
%%%
A statement has a particular truth value associated with it that is defined as the frequency of supporting evidence (denoted $f \in [0,1]$) and the confidence in the stability of that frequency (denoted $c \in [0,1]$). For example, beyond NAL-0, the statement ``Marko is a person'' is not ``100\% true'' simply because it exists. Instead, every time that aspects of Marko coincide with aspects of person, then $f$ increases. Likewise, every time aspects of Marko do not coincide with aspects of person, $f$ decreases.\footnote{The idea of ``aspects coinciding'' is formally defined in NAL, but is not discussed here for the sake of brevity. In short, a statement's $f$ is modulated by both the system's ``external'' experiences and ``internal'' reasoning---both create new evidence. See \cite{nal:wang2006} for an in depth explanation.} Thus, NAL is non-monotonic as its statement evidence can increase and decrease. To demonstrate $f$ and $c$, the above ``Marko is a person'' statement can be represented in NAL-1 as
%%%
\begin{equation*}
\ttt{lanl:marko} \rar \ttt{lanl:Person} \; <0.9,0.8>,
\end{equation*}
%%%
where, for the sake of this example, $f=0.9$ and $c=0.8$. In an RDF representation, this can be denoted

\begin{verbatim}
(lanl:marko, lanl:1234, lanl:Person)
(lanl:1234, nal:frequency, "0.9"^^xsd:float)
(lanl:1234, nal:confidence, "0.8"^^xsd:float),
\end{verbatim}
%%%
where the \ttt{lanl:1234} serves as a statement pointer allowing NAL's \ttt{nal:frequency} and \ttt{nal:confidence} constructs to reference the inheritance statement.

NAL-4 supports statements that are more analogous to the subject-object-predicate form of RDF. If Marko is denoted by the URI \ttt{lanl:marko}, Alberto by the URI \ttt{ucla:apepe}, and friendship by the URI \ttt{lanl:friend}, then in NAL-4, the statement ``Alberto is a friend of Marko'' is denoted in RDF as

\begin{verbatim}
(ucla:apepe, lanl:friend, lanl:marko).
\end{verbatim}
%%%
In NAL-4 this is represented as
%%%
\begin{equation*}
(\ttt{ucla:apepe} \times \ttt{lanl:marko}) \rar \ttt{lanl:friend} \; <0.8, 0.5>, 
\end{equation*}
%%%
where $f = 0.8$ and $c = 0.5$ are provided for the sake of the example. This statement states that the set $(\ttt{ucla:apepe}, \ttt{lanl:marko})$ inherits the property of friendship to a certain degree and stability as defined by $f$ and $c$, respectively. The RDF representation of this NAL-4 construct can be denoted

\begin{verbatim}
(lanl:2345, nal:_1, ucla:pepe)
(lanl:2345, nal:_2, lanl:marko)
(lanl:2345, lanl:3456, lanl:friend)
(lanl:3456, nal:frequency, "0.8"^^xsd:float)
(lanl:3456, nal:confidence, "0.5"^^xsd:float).
\end{verbatim}
%%%
In the triples above, \ttt{lanl:2345} serves as an set and thus, this set inherits from friendship. That is, Alberto and Marko inherit the property of friendship.

\subsubsection{Reasoning}

\begin{quote}
``In traditional logic, a `valid' or `sound' inference rule is one that never derives a \textit{false} conclusion (that is, it will be contradicted by the future experience of the system) from \textit{true} premises \cite{logic:copi1982}. [In NAL], a `valid conclusion' is one that is most consistent with the evidence in the past experience, and a `valid inference rule' is one whose conclusions are supported by the premises used to derive them.'' \cite{nal:wang2006}
\end{quote}
%%%
Given that NAL is predicated on insufficient knowledge, there is no guarantee that reasoning will produce ``true'' knowledge with respect to the world that the statements are modeling as only a subset of that world is ever known. However, this does not mean that NAL reasoning is random, instead, it is consistent with respect to what the system knows. In other words,
%%%
\begin{quote}
``the traditional definition of validity of inference rules---that is to get true conclusions from true premises---no longer makes sense in [NAL]. With insufficient knowledge and resources, even if the premises are true with respect to the past experience of the system there is no way to get infallible predictions about the future experience of the system even though the premises themselves may be challenged by new evidence.'' \cite{inherit:wang1994}
\end{quote}

The inference rules in NAL are all syllogistic in that they are based on statements sharing similar terms (i.e.~URIs) \cite{syllogism:patzig1968}. The typical inference rule in NAL has the following form
%%%
\begin{equation*}
(\tau_1 <f_1,c_1> \; \wedge \; \tau_2 <f_2,c_2>) \; \vdash \; \tau_3 <f_3, c_3>,
\end{equation*}
%%%
where $\tau_1$ and $\tau_2$ are statements that share a common term. There are four standard syllogisms used in NAL reasoning. These are enumerated below.
%%%
\begin{enumerate}\addtolength{\itemsep}{-0.5\baselineskip}
	\item deduction: $(x \rar y <f_1,c_1> \; \wedge \; y \rar z <f_2,c_2>) \; \vdash \; x \rar z <f_3, c_3>$.
	\item induction: $(x \rar y <f_1,c_1> \; \wedge \; z \rar y <f_2,c_2>) \; \vdash \; x \rar z <f_3, c_3>$.
	\item abduction: $(x \rar y <f_1,c_1> \; \wedge \; x \rar z <f_2,c_2>) \; \vdash \; y \rar z <f_3, c_3>$.
	\item exemplification: $(x \rar y <f_1,c_1> \; \wedge \; y \rar z <f_2,c_2>) \; \vdash \; z \rar x <f_3, c_3>$.
\end{enumerate}
%%%
Two other important inference rule not discussed here are choice (i.e.~what to do with contradictory evidence) and revision (i.e.~how to update existing evidence with new evidence). Each of the inference rules have a different formulas for deriving $<f_3, c_3>$ from $<f_1,c_1>$ and $<f_2,c_2>$.\footnote{Note that when the entailed statement already exists, its $<f_3, c_3>$ component is revised according to the revision rule. Revision is not discussed in this article.} These formulas are enumerated below.
%%%
\begin{enumerate}\addtolength{\itemsep}{-0.5\baselineskip}
	\item deduction: $f_3 = f_1f_2$ and $c_3 = f_1c_1f_2c_2$.
	\item induction: $f_3 = f_1$ and $c_3 = \frac{f_1c_1c_2}{f_1c_1c_2 + k}$.
	\item abduction: $f_3 = f_2$ and $c_3 = \frac{f_2c_1c_2}{f_2c_1c_2 + k}$.
	\item exemplification: $f_3 = 1$ and $c_3 = \frac{f_2c_1f_2c_2}{f_1c_1f_2c_2 + k}$.
\end{enumerate}
%%%
The variable $k \in \mbb{N}^+$ is a system specific parameter used in the determination of confidence. 

To demonstrate deduction, suppose the two statements
%%%
\begin{equation*}
\ttt{lanl:marko} \rar \ttt{lanl:Person} <0.5, 0.5>
\end{equation*}
%%%
\begin{equation*}
\ttt{lanl:Person} \rar \ttt{lanl:Mammal} <0.9,  0.9>.
\end{equation*}
%%%
Given these two statements and the inference rule for deduction, it is possible to infer
%%%
\begin{equation*}
\ttt{lanl:marko} \rar \ttt{lanl:Mammal} <0.45,  0.2025>.
\end{equation*}
%%%
Next suppose the statement
%%%
\begin{equation*}
\ttt{lanl:Dog} \rar \ttt{lanl:Mammal} <0.9,  0.9>.
\end{equation*}
%%%
Given the existing statements, induction, and a $k=1$, it is possible to infer
%%%
\begin{equation*}
\ttt{lanl:marko} \rar \ttt{lanl:Dog} <0.45,  0.0758>.
\end{equation*}
%%%
Thus, while the system is not confident, according to all that the system knows, Marko is a type of dog. This is because there are aspects of Marko that coincide with aspects of dog---they are both mammals. However, future evidence, such as fur, four legs, sloppy tongue, etc. will be further evidence that Marko and dog do not coincide and thus, the $f$ of $\ttt{lanl:marko} \rar \ttt{lanl:Dog}$ will decrease.

The significance of NAL reasoning is that all inference is based on local areas of the knowledge base. That is, all inference requires only two degrees of separation from the resource being inferred on. Moreover, reasoning is constrained by available computational resources, not by a requirement for logical completeness. Because of these two properties, the implemented reasoning system is inherently distributed and when computational resources are not available, the system does not break, it simply yields less conclusions. For the Semantic Web, it may be best to adopt a logic that is better able to take advantage of its size and inconsistency. With a reasoner that is distributable and functions under variable computational resources, and by making use of a language that is non-monotonic and supports degrees of ``truth'', NAL may serve as a more practical logic for the Semantic Web. However, this is only possible if the RDF data model is separated from the RDF semantics and NAL's subject-predicate form can be legally represented.

There are many other language constructs in NAL that are not discussed here. For an in depth review of NAL, please refer to the \textit{defacto} reference at \cite{nal:wang2006}. Moreover, for a fine discussion of the difference between logics of truth (i.e.~mathematical logic---modern predicate logic) and logics of thought (i.e.~cognitive logic---NAL), see \cite{coglogic:wang2004}.

\section{A Distributed Multi-Relational Network}\label{sec:multi-relational}

The Web of Data can be interpreted as a distributed multi-relational network---a Giant Global Graph.\footnote{The term ``graph'' is used in the mathematical domain of graph theory and the term ``network'' is used primarily in the physics and computer science domain of network theory. In this chapter, both terms are used depending on their source. Moreover, with regard to this article, these two terms are deemed synonymous with each other.} A mutli-relational network denotes a set of vertices (i.e.~nodes) that are connected to one another by set of labeled edges (i.e.~typed links).\footnote{A multi-relational network is also known as a directed labeled graph or semantic network.} In the graph and network theory community, the multi-relational network is less prevalent. The more commonly used network data structure is the single-relational network, where all edges are of the same type and thus, there is no need to label edges. Unfortunately, most network algorithms have been developed for the single-relational network data structure. However, it is possible to port all known single-relational network algorithms over to the multi-relational domain. In doing so, it is possible to leverage these algorithms on the Giant Global Graph. The purpose of this section is to 
%%%
\begin{enumerate}\addtolength{\itemsep}{-0.5\baselineskip}
	\item formalize the single-relational network (see \S \ref{sec:single-relational-networks}),
	\item formalize the multi-relational network (see \S \ref{sec:multi-relational-networks}),
	\item present a collection of common single-relational network algorithms (see \S \ref{sec:single-relational-algorithms}), and then finally,
	\item present a method for porting all known single-relational network algorithms over to the multi-relational domain (see \S \ref{sec:multi-relational-algorithms}).
\end{enumerate}

Network algorithms are useful in many respects and have been generally applied to analysis and querying. If the network models an aspect of the world, then network analysis techniques can be used to elucidate general structural properties of the network and thus, the world. Moreover, network query algorithms have been developed for searching and ranking. When these algorithms can be effectively and efficiently applied to the Giant Global Graph, the Giant Global Graph can serve as a medium for network analysis and query.

\subsection{Single-Relational Networks}\label{sec:single-relational-networks}

\begin{figure}[h!]
	\begin{center}
		\includegraphics[width=0.7\textwidth]{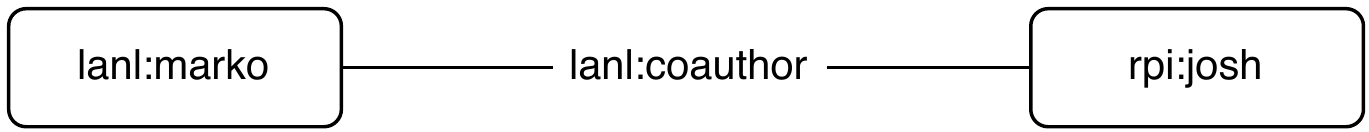}
	\caption{\label{fig:undirected} An undirected edge between two authors in an undirected single-relational network.}
	\end{center}
\end{figure}

The single-relational network represents a set of vertices that are related to one another by a homogenous set of edges. For instance, in a single-relational coauthorship network, all vertices denote authors and all edges denote a coauthoring relationship. Coauthorship exists between two authors if they have both written an article together. Moreover, coauthorship is symmetric---if person $x$ coauthored with person $y$, then person $y$ has coauthored with person $x$. In general, these types of symmetric networks are known as undirected, single-relational networks and can be denoted
%%%
\begin{equation*}
	G' = (V, E \subseteq \{V \times V\}),
\end{equation*} 
%%%
where $V$ is the set of vertices and $E$ is the set of undirected edges. The edge $\{i,j\} \in E$ states that vertex $i$ and $j$ are connected to each other. Figure \ref{fig:undirected} diagrams an undirected coauthorship edge between two author vertices.
%%%

\begin{figure}[h!]
	\begin{center}
		\includegraphics[width=0.7\textwidth]{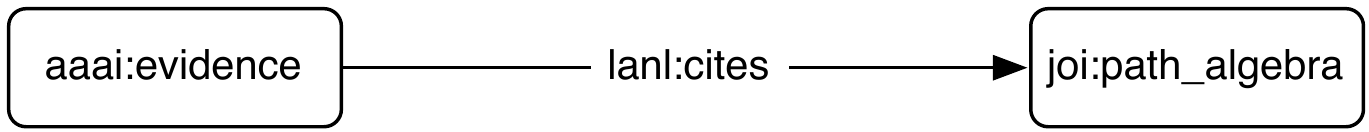}
	\caption{\label{fig:directed} A directed edge between two articles in a directed single-relational network.}
	\end{center}
\end{figure}

Single-relational networks can also be directed. For instance, in a single-relational citation network, the set of vertices denote articles and the set of edges denote citations between the articles. In this scenario, the edges are not symmetric as one article citing another does not imply that the cited article cites the citing article. Directed single-relational networks can be denoted
%%%
\begin{equation*}
	G = (V, E \subseteq (V \times V)),
\end{equation*} 
%%%
where $(i,j) \in E$ states that vertex $i$ is connected to vertex $j$. Figure \ref{fig:directed} diagrams a directed citation edge between two article vertices.
%%%

Both undirected and directed single-relational networks have a convenient matrix representation. This matrix is known as an adjacency matrix and is denoted
%%%
\begin{equation*}
	\mbf{A}_{i,j} =
		\begin{cases}
			1 & \text{if} \; (i,j) \in E \\
			0 & \text{otherwise,}
		\end{cases}
\end{equation*}
%%%
where $\mbf{A} \in \{0,1\}^{|V| \times |V|}$. If $\mbf{A}_{i,j} = 1$, then vertex $i$ is adjacent (i.e.~connected) to vertex $j$. It is important to note that there exists an information-preserving, bijective mapping between the set-theoretic and matrix representations of a network. Throughout the remainder of this section, depending on the algorithm presented, one or the other form of a network is used. Finally, note that the remainder of this section is primarily concerned with directed networks as a directed network can model an undirected network. In other words, the undirected edge $\{i,j\}$ can be represented as the two directed edges $(i,j)$ and $(j,i)$.

\subsection{Multi-Relational Networks}\label{sec:multi-relational-networks}

The multi-relational network is a more complicated structure that can be used to represent multiple types of relationships between vertices. For instance, it is possible to not only represent researchers, but also their articles in a network of edges that represent authorship, citation, etc. A directed multi-relational network can be denoted
%%%
\begin{equation*}
	M = (V, \mbb{E} = \{E_0, E_1, \ldots, E_m \subseteq (V \times V)\}),
\end{equation*} 
%%%
where $\mbb{E}$ is a family of edge sets such that any $E_k \in \mbb{E} : 1 \leq k \leq m$ is a set of edges with a particular meaning (e.g.~authorship, citation, etc.). A multi-relational network can be interpreted as a collection of single-relational networks that all share the same vertex set. Another representation of a multi-relational network is similar to the one commonly employed to define an RDF graph. This representation is denoted
%%%
\begin{equation*}
	M' \subseteq (V \times \Omega \times V),
\end{equation*} 
%%%
where $\Omega$ is the set of edge labels. In this representation if $i,j \in V$ and $k \in \Omega$, then the triple $(i,k,j)$ states that vertex $i$ is connected to vertex $j$ by the relationship type $k$.

\begin{figure}[h!]
	\begin{center}
		\includegraphics[width=0.7\textwidth]{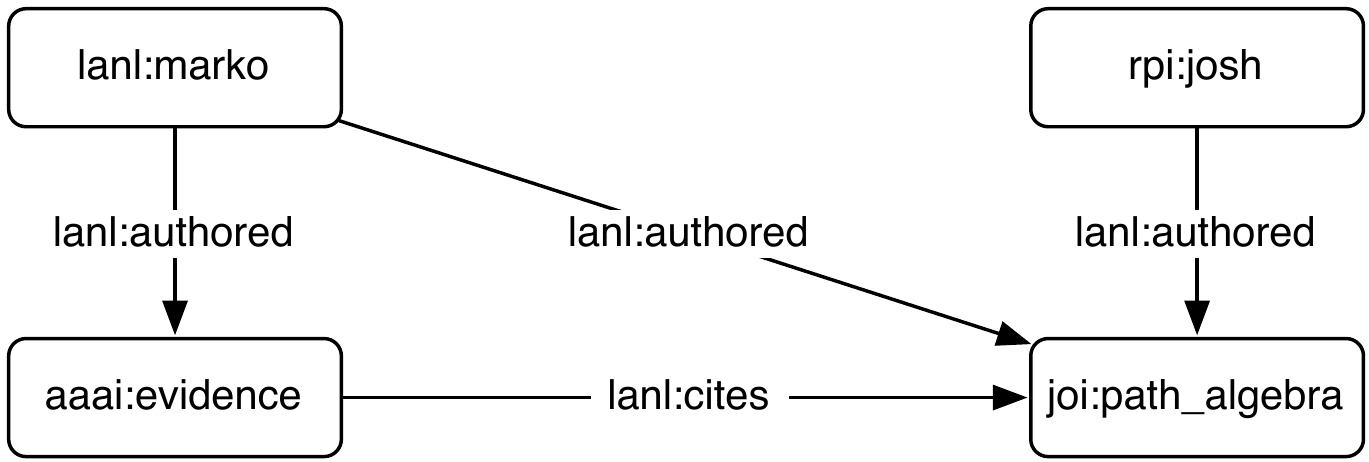}
	\caption{\label{fig:multi-relational} Multiple types of edges between articles and scholars in a directed multi-relational network.}
	\end{center}
\end{figure}

Figure \ref{fig:multi-relational} diagrams multiple relationship types between scholars and articles in a multi-relational network.
%%%

Like the single-relational network and its accompanying adjacency matrix, the multi-relational network has a convenient $3$-way tensor representation. This $3$-way tensor is denoted
%%%
\begin{equation*}
	\mca{A}^k_{i,j} =
		\begin{cases}
			1 & \text{if} \; (i,j) \in E_k : 1 \leq k \leq m \\
			0 & \text{otherwise.}
		\end{cases}
\end{equation*}
%%%
This representation can be interpreted as a collection of adjacency matrix ``slices", where each slice is a particular edge type. In other words, if $\mca{A}^k_{i,j} = 1$, then $(i,k,j) \in M'$. Like the relationship between the set-theoretic and matrix forms of a single-relational network, $M$, $M'$, and $\mca{A}$ can all be mapped onto one another without loss of information. Each representation will be used depending on the usefulness of its form with respect to the idea being expressed.

On the Giant Global Graph, RDF serves as the specification for graphing resources. Vertices are denoted by URIs, blank nodes, and literals and the edge labels are denoted by URIs. Multi-relational network algorithms can be used to exploit the Giant Global Graph. However, there are few algorithms dedicated specifically to multi-relational networks. Most network algorithms have been designed for single-relational networks. The remainder of this section will discuss some of the more popular single-relational network algorithms and then present a method for porting these algorithms (as well as other single-relational network algorithms) over to the multi-relational domain. This section concludes with a distributable and scalable method for executing network algorithms on the Giant Global Graph.

\subsection{Single-Relational Network Algorithms}\label{sec:single-relational-algorithms}

The design and study of graph and network algorithms is conducted primarily by mathematicians (graph theory) \cite{graph:chartrand1977}, physicists and computer scientists (network theory) \cite{netanal:brandes2005}, and social scientists (social network analysis) \cite{socialanal:wasserman1994}. Many of the algorithms developed in these domains can be used together and form the general-purpose ``toolkit" for researchers doing network analysis and for engineers developing network-based services. The following itemized list presents a collection of the single-relational network algorithms that will be reviewed in this subsection. As denoted with its name in the itemization, each algorithm can be used to identify properties of vertices, paths, or the network. Vertex metrics assign a real value to a vertex. Path metrics assign a real value to a path. And finally, network metrics assign a real value to the network as a whole.
%%%
\begin{itemize}\addtolength{\itemsep}{-0.5\baselineskip}
	\item shortest path: path metric (\S \ref{sec:shortpath})
	\item eccentricity: vertex metric (\S \ref{sec:ecenraddiam})
	\item radius: network metric (\S \ref{sec:ecenraddiam})
	\item diameter: network metric (\S \ref{sec:ecenraddiam})
	\item closeness: vertex metric (\S \ref{sec:closebetween})
	\item betweenness: vertex metric (\S \ref{sec:closebetween})
	\item stationary probability distribution: vertex metric (\S \ref{sec:stationary})
	\item PageRank: vertex metric (\S \ref{sec:pagerank})
	\item spreading activation: vertex metric (\S \ref{sec:spread})
	\item assortative mixing: network metric (\S \ref{sec:assortativity})
\end{itemize}

A simple intuitive approach to determine the appropriate algorithm to use for an application scenario is presented in \cite{advance:kosch2004}. In short, various factors come into play when selecting a network algorithm such as the topological features of the network (e.g.~its connectivity and its size), the computational requirements of the algorithms (e.g.~its complexity), the type of results that are desired (e.g.~personalized or global), and the meaning of the algorithm's result (e.g.~geodesic-based, flow-based, etc.). The following sections will point out which features describe the presented algorithms.

\subsubsection{Shortest Path}\label{sec:shortpath}

The shortest path metric is the foundation of all other geodesic metrics. The other geodesic metrics discussed are eccentricity, radius, diameter, closeness, and betweenness. A shortest path is defined for any two vertices $i,j \in V$ such that the sink vertex $j$ is reachable from the source vertex $i$. If $j$ is unreachable from $i$, then the shortest path between $i$ and $j$ is undefined. Thus, for geodesic metrics, it is important to only considered strongly connected networks, or strongly connected components of a network.\footnote{Do not confuse a strongly connected network with a fully connected network. A fully connected network is where every vertex is connected to every other vertex directly. A strongly connected network is where every vertex is connected to every other vertex indirectly (i.e.~there exists a path from any $i$ to any $j$).} The shortest path between any two vertices $i$ and $j$ in a single-relational network is the smallest of the set of all paths between $i$ and $j$. If $\rho : V \times V \rar Q$ is a function that takes two vertices and returns the set of all paths $Q$ where for any $q \in Q$, $q = (i,\dots,j)$, then the length of the shortest path between $i$ and $j$ is $min(\bigcup_{q \in Q} |q|-1)$, where $min$ returns the smallest value of its domain. The shortest path function is denoted $s : V \times V \rar \mbb{N}$ with the function rule
%%%
\begin{equation*}
	s(i,j) = min\left(\bigcup_{q \in \rho(i,j)} |q|-1 \right).
\end{equation*}

There are many algorithms to determine the shortest path between vertices in a network. Dijkstra's method is perhaps the most popular as it is the typical algorithm taught in introductory algorithms classes \cite{short:dijkstra1959}. However, if the network is unweighted, then a simple breadth-first search is a more efficient way to determine the shortest path between $i$ and $j$. Starting from $i$ a ``fan-out'' search for $j$ is executed where at each time step, adjacent vertices are traversed to. The first path that reaches $j$ is the shortest path from $i$ to $j$.

\subsubsection{Eccentricity, Radius, and Diameter}\label{sec:ecenraddiam}

The radius and diameter of a network require the determination of the eccentricity of every vertex in $V$. The eccentricity of a vertex $i$ is the largest shortest path between $i$ and all other vertices in $V$ such that the eccentricity function $e : V \rar \mbb{N}$ has the rule
%%%
\begin{equation*}
	e(i) = max\left(\bigcup_{j \in V} s(i,j) : i \neq j \right),
\end{equation*}
%%%	
where $max$ returns the largest value of its domain \cite{eccen:harary1995}. The eccentricity metric calculates $|V| - 1$ shortest paths of a particular vertex.

The radius of the network is the minimum eccentricity of all vertices in $V$ \cite{socialanal:wasserman1994}. The function $r : G \rar \mbb{N}$ has the rule 
%%%
\begin{equation*}
	r(G) = min\left(\bigcup_{i \in V} e(i) \right).
\end{equation*}

Finally, the diameter of a network is the maximum eccentricity of the vertices in $V$ \cite{socialanal:wasserman1994}. The function $d : G \rar \mbb{N}$ has the rule 
%%%
\begin{equation*}
	d(G) = max\left(\bigcup_{i \in V} e(i) \right).
\end{equation*}

The diameter of a network is, in some cases, telling of the growth properties of the network (i.e.~the general principle by which new vertices and edges are added). For instance, if the network is randomly generated (edges are randomly assigned between vertices), then the diameter of the network is much larger then if the network is generated according to a more ``natural growth'' function such as a preferential attachment model, where highly connected vertices tend to get more edges (colloquially captured by the phrase ``the rich get richer'') \cite{diamscale:2004}. Thus, in general, natural networks tend to have a much smaller diameter. This was evinced by an empirical study of the World Wide Web citation network, where the diameter of the network was concluded to be only $19$ \cite{diamwww:albert1999}.

\subsubsection{Closeness and Betweenness Centrality}\label{sec:closebetween}

Closeness and betweenness centrality are popular network metrics for determining the ``centralness'' of a vertex and have been used in sociology \cite{socialanal:wasserman1994}, bioinformatics \cite{gene:ozgur2008}, and bibliometrics \cite{bollen:mesur2008}. Centrality is a loose term that describes the intuitive notion that some vertices are more connected/integral/central/influential within the network than others. Closeness centrality is one such centrality measure and is defined as the mean shortest path between some vertex $i$ and all the other vertices in $V$ \cite{close:bavelas1950,close:leavitt1951,close:sabaidussi1966}. The function $c : V \rar \mbb{R}$ has the rule
%%%
\begin{equation*}
	c(i) = \frac{1}{\sum_{j \in V} s(i,j)}.
\end{equation*}

Betweenness centrality is defined for a vertex in $V$ \cite{between:freeman1977,betweeness:brandes2001}. The betweenness of $i \in V$ is the number of shortest paths that exist between all vertices $j,k \in V$ that have $i$ in their path divided by the total number of shortest paths between $j$ and $k$, where $i \neq j \neq k$. If $\sigma : V \times V \rar Q$ is the function that returns the set of shortest paths between any two vertices $j$ and $k$ such that 
%%%
\begin{equation*}
	\sigma(j,k) = \bigcup_{q \in p(j,k)} q : |q|-1 = s(j,k)
\end{equation*}
%%%
and $\hat{\sigma} : V \times V \times V \rar Q$ is the set of shortest paths between two vertices $j$ and $k$ that have $i$ in the path, where
%%%
\begin{equation*}
	\hat{\sigma}(j,k,i) = \bigcup_{q \in p(j,k)} q : (|q|-1 = s(j,k) \; \wedge \; i \in q), 
\end{equation*}
%%%
then the betweenness function $b : V \rar \mbb{R}$ has the rule
%%%
\begin{equation*}
	b(i) = \sum_{i \neq j \neq k \in V} \frac{|\hat{\sigma}(j,k,i)|}{|\sigma(j,k)|}.
\end{equation*}

There are many variations to the standard representations presented above. For a more in depth review on these metrics, see \cite{socialanal:wasserman1994} and \cite{netanal:brandes2005}. Finally, centrality is not restricted only to geodesic metrics. The next three algorithms are centrality metrics based on random walks or ``flows'' through a network.

\subsubsection{Stationary Probability Distribution}\label{sec:stationary}

A Markov chain is used to model the states of a system and the probability of transition between states \cite{markov:haggstrom2002}. A Markov chain is best represented by a probabilistic, single-relational network where the states are vertices, the edges are transitions, and the edge weights denote the probability of transition. A probabilistic, single-relational network can be denoted 
%%%
\begin{equation*}
G'' = \left(V, E \subseteq (V \times V), \omega : E \rar [0,1]\right)
\end{equation*}
%%%
where $\omega$ is a function that maps each edge in $E$ to a probability value. The outgoing edges of any vertex form a probability distribution that sums to $1.0$. In this section, all outgoing probabilities from a particular vertex are assumed to be equal. Thus, $ \forall j,k \in \Gamma^+(i) :  \omega(i,j) = \omega(i,k)$, where $\Gamma^+(i) \subseteq V$ is the set of vertices adjacent to $i$.

A random walker is a useful way to visualize the transitioning between vertices. A random walker is a discrete element that exists at a particular $i \in V$ at a particular point in time $t \in \mathbb{N}^+$. If the vertex at time $t$ is $i$ then the next vertex at time $t+1$ will be one of the vertices adjacent to $i$ in $\Gamma^+(i)$. In this manner, the random walker makes a probabilistic jump to a new vertex at every time step. As time $t$ goes to infinity a unique stationary probability distribution emerges if and only if the network is aperiodic and strongly connected. The stationary probability distribution expresses the probability that the random walker will be at a particular vertex in the network. In matrix form, the stationary probability distribution is represented by a row vector $\pi \in [0,1]^{|V|}$, where $\pi_i$ is the probability that the random walker is at $i$ and $\sum_{i \in V} \vpi_i = 1.0$. If the network is represented by the row-stochastic adjacency matrix
%%%
\begin{equation*}
	\mbf{A}_{i,j} =
		\begin{cases}
			\frac{1}{|\Gamma^+(i)|}  & \text{if } (i,j) \in E \\
			0 & \text{otherwise}
		\end{cases}
\end{equation*}
%%%
and if the network is aperiodic and strongly connected, then there exists some $\pi$ such that $\pi\mbf{A} = \pi$. Thus, the stationary probability distribution is the primary eigenvector of $\mbf{A}$. The primary eigenvector of a network is useful in ranking its vertices as those vertices that are more central are those that have a higher probability in $\pi$. Thus, intuitively, where the random walker is likely to be is a indicator of how central the vertex is. However, if the network is not strongly connected (very likely for most natural networks), then a stationary probability distribution does not exist.

\subsubsection{PageRank}\label{sec:pagerank}

PageRank makes use of the random walker model previously presented \cite{anatom:brin1998}. However, in PageRank, the random walker does not simply traverse the single-relational network by moving between adjacent vertices, but instead has a probability of jumping, or ``teleporting'', to some random vertex in the network. In some instances, the random walker will follow an outgoing edge from its current vertex location. In other instances, the random walker will jump to some other random vertex in the network that is not necessarily adjacent to it. The benefit of this model is that it ensures that the network is strongly connected and aperiodic and thus, there exists a stationary probability distribution. In order to calculate PageRank, two networks are used. The standard single-relational network is represented as the row-stochastic adjacency matrix
%%%
\begin{equation*}
	\mbf{A}_{i,j} =
		\begin{cases}
			\frac{1}{|\Gamma^+(i)|}  & \text{if } (i,j) \in E \\
			\frac{1}{|V|} & \text{otherwise.}
		\end{cases}
\end{equation*}
%%%
Any $i \in V$ where $\Gamma^+(i) = \emptyset$ is called a ``rank-sink". Rank-sinks ensure that the network is not strongly connected. To rectify this connectivity problem, all vertices that are rank-sinks are connected to every other vertex with probability $\frac{1}{|V|}$. Next, for teleportation, a fully connected network is created that is denoted $\mbf{B}_{i,j} = \frac{1}{|V|}$. 

The random walker will choose to use $\mbf{A}$ or $\mbf{B}$ at time step $t$ as its transition network depending on the probability value $\alpha \in (0,1]$, where in practice, $\alpha=0.85$. This means that 85\% of the time the random walker will use the edges in $\mbf{A}$ to traverse, and the other 15\% of the time, the random walker will use the edges in $\mbf{B}$. The $\alpha$-biased union of the networks $\mbf{A}$ and $\mbf{B}$ guarantees that the random walker is traversing an strongly connected and aperiodic network. The random walker's traversal network can be expressed by the matrix
%%%
\begin{equation*}
	\mbf{C} = \alpha\mbf{A} + (1-\alpha)\mbf{B}.
\end{equation*}

The PageRank row vector $\pi \in [0,1]^{|V|}$ has the property $\pi\mbf{C} = \pi$. Thus, the PageRank vector is the primary eigenvector of the modified single-relational network. Moreover, $\pi$ is the stationary probability distribution of $\mbf{C}$. From a certain perspective, the primary contribution of the PageRank algorithm is not in the way it is calculated, but in how the network is modified to support a convergence to a stationary probability distribution. PageRank has been popularized by the Google search engine and has been used as a ranking algorithm in various domains. Relative to the geodesic centrality algorithms presented previous, PageRank is a more efficient way to determine a centrality score for all vertices in a network. However, calculating the stationary probability distribution of a network is not cheap and for large networks, can not be accomplished in real-time. Local rank algorithms are more useful for real-time results in large-scale networks such as the Giant Global Graph.

\subsubsection{Spreading Activation}\label{sec:spread}

Both the stationary probability distribution and PageRank are global rank metrics. That is, they rank all vertices relative to all vertices and as such, require a full network perspective. However, for many applications, a local rank metric is desired. Local rank metrics rank a subset of vertices relative to some set of source vertices. Local rank metrics have the benefit of being faster to compute and being relative to a particular area of the network. For large-scale networks, local rank metrics are generally more practical for real-time queries.

Perhaps the most popular local rank metric is spreading activation. Spreading activation is a network analysis technique that was inspired by the spreading activation potential found in biological neural networks \cite{spread:collins1975,spread:anderson1983,neural:haykin1999}. This algorithm (and its many variants) has been used extensively in semantic network reasoning and recommender systems. The purpose of the algorithm is to expose, in a computationally efficient manner, those vertices which are closest (in terms of a flow distance) to a particular set of vertices. For example, given $i,j,k \in V$, if there exists many short recurrent paths between vertex $i$ and vertex $j$ and not so between $i$ and $k$, then it can be assumed that vertex $i$ is more ``similar'' to vertex $j$ than $k$. Thus, the returned ranking will rank $j$ higher than $k$ relative to $i$. In order to calculate this distance, ``energy'' is assigned to vertex $i$. Let $x \in [0,1]^{|V|}$ denote the energy vector, where at the first time step all energy is at $i$ such that $x_i^1 = 1.0$. The energy vector is propagated over $\mbf{A}$ for $\hat{t} \in \mbb{N}^+$ number of steps by the equation $x^{t+1} = x^{t}\mbf{A} : t+1 \leq \hat{t}$. Moreover, at every time step, $x$ is decayed some amount by $\delta \in [0,1]$. At the end of the process, the vertex that had the most energy flow through it (as recorded by $\pi \in \mbb{R}^{|V|}$) is considered the vertex that is most related to vertex $i$. Algorithm \ref{alg:spread} presents this spreading activation algorithm. The resultant $\pi$ provides a ranking of all vertices at most $\hat{t}$ steps away from $i$.
%%%
\restylealgo{boxed}
\begin{algorithm}[h!]
\dontprintsemicolon
\Begin{
$t = 1$ \;
\While{$t \leq \hat{t}$} {
	$\vpi = \vpi + x$ \;
	$x = (\delta x)\mbf{A}$ \;
	$t = t + 1$ \;
}
\Return{$\vpi$}
}
\caption{\label{alg:spread}A spreading activation algorithm.}
\end{algorithm}

A class of algorithms known as ``priors'' algorithms perform computations similar to the local rank spreading activation algorithm, but do so using a stationary probability distribution \cite{markov:white2003}. Much like the PageRank algorithm distorts the original network, priors algorithms distort the local neighborhood of the graph and require at every time step, with some probability, that all random walkers return to their source vertex. The long run behavior of such systems yield a ranking biased towards (or relative to) the source vertices and thus, can be characterized as local rank metrics.

\subsubsection{Assortative Mixing}\label{sec:assortativity}

The final single-relational network algorithm discussed is assortative mixing. Assortative mixing is a network metric that determines if a network is assortative (colloquially captured by the phrase ``birds of a feather flock together"), disassortative (colloquially captured by the phrase ``opposites attract"), or uncorrelated. An assortative mixing algorithm returns values in $[-1,1]$, where $1$ is assortative, $-1$ is disassortative, and $0$ is uncorrelated. Given a collection of vertices and metadata about each vertex, it is possible to determine the assortative mixing of the network. There are two assortative mixing algorithms: one for scalar or numeric metadata (e.g.~age, weight, etc.) and one for nominal or categorical metadata (e.g.~occupation, sex, etc.). In general, an assortative mixing algorithm can be used to answer questions such as:
%%%
\begin{itemize}\addtolength{\itemsep}{-0.5\baselineskip}
	\item Do friends in a social network tend to be the same age?
	\item Do colleagues in a coauthorship network tend to be from the same university?
	\item Do relatives in a kinship network tend to like the same foods? 
\end{itemize}
%%%
Note that to calculate the assortative mixing of a network, vertices must have metadata properties. The typical single-relational network $G = (V,E)$ does not capture this information. Therefore, assume some other data structure that stores metadata about each vertex.

The original publication defining the assortative mixing metric for scalar properties used the parametric Pearson correlation of two vectors \cite{newman:assort}.\footnote{Note that for metadata property distributions that are not normally distributed, a non-parametric correlation such as the Spearman $\rho$ or Kendall $\tau$ may be the more useful correlation coefficient.} One vector is the scalar value of the vertex property for the vertices on the tail of all edges. The other vector is the scalar value of the vertex property for the vertices on the head of all the edges. Thus, the length of both vectors is $|E|$ (i.e.~the total number of edges in the network).  Formally, the Pearson correlation-based assortativity is defined as
%%%
\begin{equation*}
r = \frac{|E| \sum_{i} j_i k_i -  \sum_i j_i \sum_i  k_i}{\sqrt{\left[|E| \sum_i j^2_i - \left(\sum_i j_i\right)^2\right]\left[|E| \sum_i k^2_i -\left(\sum_i k_i\right)^2\right]}},
\end{equation*}
%%%
where $j_i$ is the scalar value of the vertex on the tail of edge $i$, and $k_i$ is the scalar value of the vertex on the head of edge $i$. For nominal metadata, the equation
%%%
\begin{equation*}
r = \frac{\sum_{p} e_{pp} - \sum_p a_p b_p}{1 - \sum_p a_p b_p}
\end{equation*}
%%%
yields a value in $[-1,1]$ as well, where $e_{pp}$ is the number of edges in the network that have property value $p$ on both ends, $a_p$ is the number of edges in the network that have property value $p$ on their tail vertex, and $b_p$ is the number of edges that have property value $p$ on their head vertex \cite{newman:mixpatt2003}.

\subsection{Porting Algorithms to the Multi-Relational Domain}\label{sec:multi-relational-algorithms}

All the aforementioned algorithms are intended for single-relational networks. However, it is possible to map these algorithms over to the multi-relational domain and thus, apply them to the Giant Global Graph. In the most simple form, it is possible to ignore edge labels and simply treat all edges in a multi-relational network as being equal. This method, of course, does not take advantage of the rich structured data that multi-relational networks offer. If only a particular single-relational slice of the multi-relational network is desired (e.g.~a citation network, \ttt{lanl:cites}), then this single-relational component can be isolated and subjected the previously presented single-relational network algorithms. However, if a multi-relational network is to be generally useful, then a method that takes advantage of the various types of edges in the network is desired. The methods presented next define abstract/implicit paths through a network. By doing so, a multi-relational network can be redefined as a ``semantically rich'' single-relational network. For example, in Figure \ref{fig:multi-relational}, there does not exist \ttt{lanl:authorCites} edges (i.e.~if person $i$ wrote an article that cites the article of person $j$, then it is true that $i$ \ttt{lanl:authorCites} $j$). However, this edge can be automatically generated by making use of the \ttt{lanl:authored} and \ttt{lanl:cites} edges. In this way, a breadth-first search or a random walk can use these automatically generated, semantically rich edges. By using generated edges, it is possible to treat a multi-type subset of the multi-relational network as a single-relational network.

\subsubsection{A Multi-Relational Path Algebra}

A path algebra is presented to map a multi-relational network to a single-relational network in order to expose the multi-relational network to single-relational network algorithms. The multi-relational path algebra summarized is discussed at length in \cite{pathalg:rodriguez2008}. In short, the path algebra manipulates a multi-relational tensor, $\mca{A} \in \{0,1\}^{|V| \times |V| \times |\mbb{E}|}$, in order to derive a semantically-rich, weighted single-relational adjacency matrix, $\mbf{A} \in \mbb{R}^{|V| \times |V|}$. Uses of the algebra can be generally defined as
%%%
\begin{equation*}
	\Delta: \{0,1\}^{|V| \times |V| \times |\mbb{E}|} \rar \mbb{R}^{|V| \times |V|},
\end{equation*}
%%%
where $\Delta$ is the path operation defined.

There are two primary operations used in the path algebra: traverse and filter.\footnote{Other operations not discussed in this section are merge and weight. For a in depth presentation of the multi-relational path algebra, see \cite{pathalg:rodriguez2008}.} The traverse operation is denoted $\cdot: \mbb{R}^{|V| \times |V|} \times \mbb{R}^{|V| \times |V|}$ and uses standard matrix multiplication as its function rule. Traverse is used to ``walk'' the multi-relational network. The idea behind traverse is first described using a single-relational network example. If a single-relational adjacency matrix is raised to the second power (i.e.~multiplied with itself) then the resultant matrix denotes how many paths of length $2$ exist between vertices \cite{graph:chartrand1977}. That is, $\mbf{A}^{(2)}_{i,j}$ (i.e.~$(\mbf{A} \cdot \mbf{A})_{i,j}$) denotes how many paths of length $2$ go from vertex $i$ to vertex $j$. In general, for any power $p$,
%%%
\begin{equation*}
	\mbf{A}_{i,j}^{(p)} =  \sum_{l \in V} \mbf{A}_{i,l}^{(p-1)} \cdot \mbf{A}_{l,j} : p \geq 2.
\end{equation*}
%%%
This property can be applied to a multi-relational tensor. If $\mca{A}^1$ and $\mca{A}^2$ are multiplied together then the result adjacency matrix denotes the number of paths of type $1\rar 2$ that exist between vertices. For example, if $\mca{A}^1$ is the coauthorship adjacency matrix, then the adjacency matrix $\mbf{Z} = \mca{A}^1\cdot {\mca{A}^1}^\top$ denotes how many coauthorship paths exist between vertices, where $\top$ transposes the matrix (i.e.~inverts the edge directionality). In other words if Marko (vertex $i$) and Johan (vertex $j$) have written $19$ papers together, then $\mbf{Z}_{i,j} = 19$.  However, given that the identity element $\mbf{Z}_{i,i}$ may be greater than $0$ (i.e.~a person has coauthored with themselves), it is important to remove all such reflexive coauthoring paths back to the original author. In order to do this, the filter operation is used. Given the identify matrix $\mbf{I}$ and the all $1$ matrix $\mbf{1}$,
%%%
\begin{equation*}
	\mbf{Z} = \left(\mca{A}^1\cdot {\mca{A}^1}^\top \right) \circ \left(\mbf{1} - \mbf{I} \right), 
\end{equation*}
%%%%
yields a true coauthorship adjacency matrix, where $\circ: \mbb{R}^{|V| \times |V|} \times \mbb{R}^{|V| \times |V|}$ is the entry-wise Hadamard matrix multiplication operation \cite{matrix:horn1994}. Hadamard matrix multiplication is defined as
%%%
\begin{equation*}
	\mbf{A} \circ \mbf{B} = \left[
	\begin{array}{ccc}
		\mbf{A}_{1,1} \cdot \mbf{B}_{1,1} & \cdots & \mbf{A}_{1,j} \cdot \mbf{B}_{1,m} \\
		\vdots & \ddots & \vdots \\
		\mbf{A}_{n,1} \cdot \mbf{B}_{n,1} & \cdots & \mbf{A}_{n,m} \cdot \mbf{B}_{n,m} \\
	\end{array} \right ] .
\end{equation*}
%%%
In this example, the Hadamard entry-wise multiplication operation applies an ``identify filter'' to $\left(\mca{A}^1\cdot {\mca{A}^1}^\top \right)$ that removes all paths back to the source vertices (i.e.~back to the identity vertices) as it sets $\mbf{Z}_{i,i} = 0$. This example demonstrates that a multi-relational network can be mapped to a semantically-rich, single-relational network. In the original multi-relational network, there exists no coauthoring relationship. However, this relation exists implicitly by means of traversing and filtering particular paths.\footnote{While not explored in \cite{pathalg:rodriguez2008}, it is possible to use the path algebra to create inference rules in a manner analogous to the Semantic Web Rule Language (SWRL) \cite{swrl:horrocks2004}. Moreover, as explored in \cite{pathalg:rodriguez2008}, it is possible to perform any arbitrary SPARQL query \cite{sparql:prud2004} using the path algebra (save for greater-than/less-than comparisons of and regular expressions on literals).}

The benefit of the summarized path algebra is that is can express various abstract paths through a multi-relational tensor in an algebraic form. Thus, given the theorems of the algebra, it is possible to simplify expressions in order to derive more computationally efficient paths for deriving the same information. The primary drawback of the algebra is that it is a matrix algebra that globally operates on adjacency matrix slices of the multi-relational tensor $\mca{A}$. Given that size of the Giant Global Graph, it is not practical to execute global matrix operations. However, these path expressions can be used as an abstract path that a discrete ``walker'' can take when traversing local areas of the graph. This idea is presented next.

\subsubsection{Multi-Relational Grammar Walkers}

Previously, both the stationary probability distribution, PageRank, and spreading activation were defined as matrix operations. However, it is possible to represent these algorithms using discrete random walkers. In fact, in many cases, this is the more natural representation both in terms of intelligibility and scalability. For many, it is more intuitive to think of these algorithms as being executed by a discrete random walker moving from vertex to vertex recording the number of times it has traversed each vertex. In terms of scalability, all of these algorithms can be approximated by using less walkers and thus, less computational resources. Moreover, when represented as a swarm of discrete walkers, the algorithm is inherently distributed as a walker is only aware of its current vertex and those vertices adjacent to it.

For multi-relational networks, this same principle applies. However, instead of randomly choosing an adjacent vertex to traverse to, the walker chooses a vertex that is dependent upon an abstract path description defined for the walker. Walkers of this form are called grammar-based random walkers \cite{grammar:rodriguez2008}. A path for a walker can be defined using any language such as the path algebra presented previous or SPARQL \cite{sparql:prud2004}. The following examples are provided in SPARQL as it is the \textit{defacto} query language for the Web of Data. Given the coauthorship path description
%%%
\begin{equation*}
	\left(\mca{A}^1\cdot {\mca{A}^1}^\top \right) \circ \left(\mbf{1} - \mbf{I} \right), 
\end{equation*}
%%%
it is possible to denote this as a local walker computation in SPARQL as

\begin{verbatim}
SELECT ?dest WHERE {
  @ lanl:authored ?x .
  ?dest lanl:authored ?x .
  FILTER (@ != ?dest)	
}
\end{verbatim}
%%%
where the symbol \ttt{@} denotes the current location of the walker (i.e.~a parameter to the query) and \ttt{?dest} is a collection of potential locations for the walker to move to (i.e.~the return set of the query). It is important to note that the path algebra expression performs a global computation while the SPARQL query representation distributes the computation to the individual vertices (and thus, individual walkers). Given the set of resources that bind to \ttt{?dest}, the walker selects a single resource from that set and traverses to it. At which point, \ttt{@} is updated to that selected resource value. This process continues indefinitely and, in the long run behavior, the walker's location probability over $V$ denotes the stationary distribution of the walker in the Giant Global Graph according to the abstract coauthorship path description. The SPARQL query redefines what is meant by an adjacent vertex by allowing longer paths to be represented as single edges. Again, this is why it is stated that such mechanisms yield semantically rich, single-relational networks.

\begin{figure}[h!]
	\begin{center}
		\includegraphics[width=0.675\textwidth]{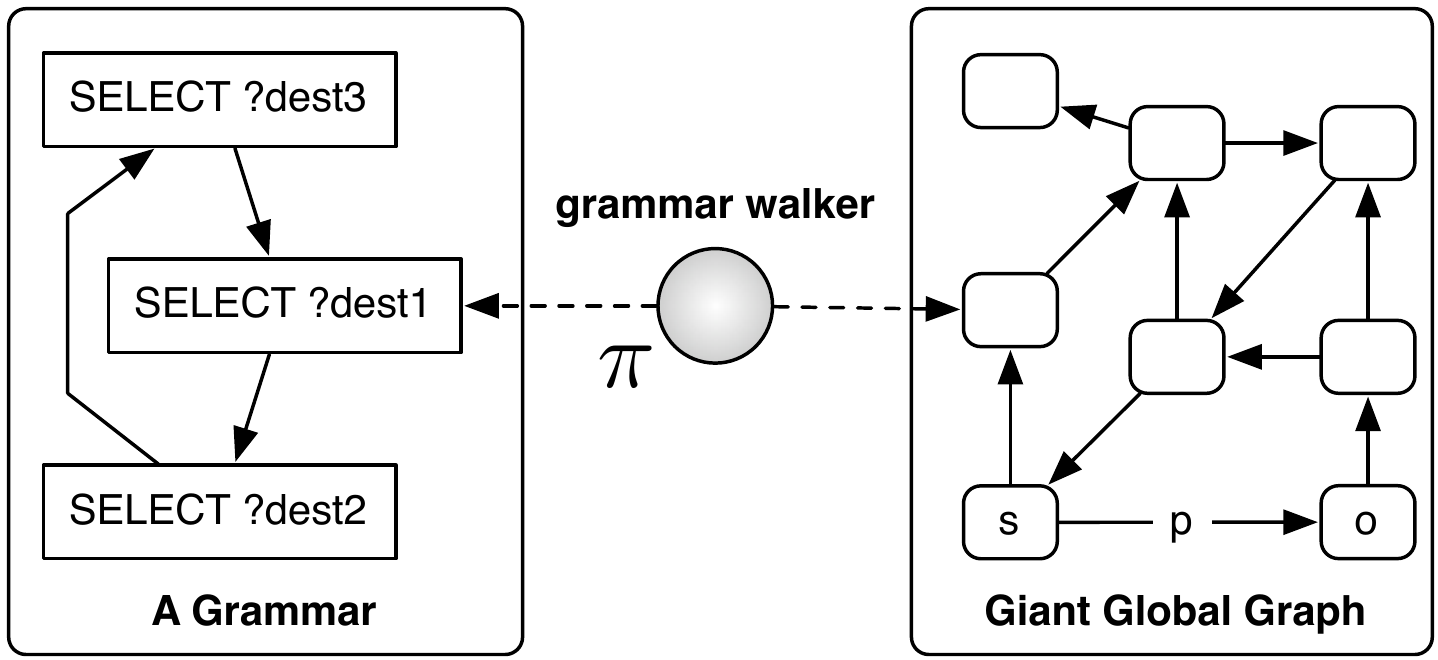}
	\caption{\label{fig:grammar-walk}A grammar walker maintains its state in the Giant Global Graph (its current vertex location) and its state in the grammar (its current grammar location---SPARQL query). After executing its current SPARQL query, the walker moves to a new vertex in the Giant Global Graph as well as to a new grammar location in the grammar.}
	\end{center}
\end{figure}

In the previous coauthorship example, the grammar walker, at every vertex it encounters, executes the same SPARQL query to locate ``adjacent'' vertices. In more complex grammars, it is possible to chain together SPARQL queries into a graph of expressions such that the walker moves not only through the Giant Global Graph, but also through a web of SPARQL queries. Each SPARQL query defines a different abstract edge to be traversed. This idea is diagrammed in Figure \ref{fig:grammar-walk}, where the grammar walker ``walks'' both the grammar and the Giant Global Graph.
%%%

To demonstrate a multiple SPARQL query grammar, a PageRank coauthorship grammar is defined using two queries. The first query was defined above and the second query is

\begin{verbatim}
SELECT ?dest WHERE {
  ?dest rdf:type lanl:Person
}
\end{verbatim}
%%%
This rule serves as the ``teleportation'' function utilized in PageRank to ensure a strongly connected network. Thus, if there is a $\alpha$ probability that the first query will be executed and a $(1-\alpha)$ probability that the second rule will be executed, then coauthorship PageRank in the Giant Global Graph is computed. Of course, the second rule can be computationally expensive, but it serves to elucidate the idea.\footnote{Note that this description is not completely accurate as ``rank sinks'' in the first query (when $\ttt{?dest} = \emptyset$) will halt the process. Thus, in such cases, when the process halts, the second query should be executed. At which point, rank sinks are alleviated and PageRank is calculated.} It is noted that the stationary probability distribution and the PageRank of the Giant Global Graph can be very expensive to compute if the grammar does not reduce the traverse space to some small subset of the full Giant Global Graph. In many cases, grammar walkers are more useful for calculating semantically meaningful spreading activations. In this form, the Giant Global Graph can be searched efficiently from a set of seed resources and a set of walkers that do not iterate indefinitely, but instead, for some finite number of steps.

The geodesic algorithms previously defined in \S \ref{sec:single-relational-algorithms} can be executed in an analogous fashion using grammar-based geodesic walkers \cite{geodesics:rodriguez2007}. The difference between a geodesic walker and a random walker is that the geodesic walker creates a ``clone'' walker each time it is adjacent to multiple vertices. This is contrasted to the random walker, where the random walker randomly chooses a single adjacent vertex. This cloning process implements a breadth-first search. It is noted that geodesic algorithms have high algorithmic complexity and thus, unless the grammar can be defined such that only a small subset of the Giant Global Graph is traversed, then such algorithms should be avoided. In general, the computational requirements of the algorithms in single-relational networks also apply to multi-relational networks. However, in multi-relational networks, given that adjacency is determined through queries, multi-relational versions of these algorithms are more costly. Given that the Giant Global Graph will soon grow to become the largest network instantiation in existence, being aware of such computational requirements is a necessary. 

Finally, a major concern with the Web of Data as it is right now is that data is pulled to a machine for processing. That is, by resolving an \ttt{http}-based URI, an RDF subgraph is returned to the retrieving machine. This is the method advocated by the Linked Data community \cite{linkeddata:bizer2008}. Thus, walking the Giant Global Graph requires pulling large amounts of data over the wire. For large network traversals, instead of moving the data to the process, it may be better to move the process to the data. By discretizing the process (e.g.~using walkers) it is possible to migrate walkers between the various servers that support the Giant Global Graph. These ideas are being further developed in future work. 

\section{A Distributed Object Repository}\label{sec:object-repository}

The Web of Data can be interpreted as a distributed object repository---a Web of Process. An object, from the perspective of object-oriented programming, is defined as a discrete entity that maintains
%%%
\begin{itemize}
	\item fields: properties associated with the object. These may be pointers to literal primitives such as characters, integers, etc. or pointers to other objects.
	\item methods: behaviors associated with the object. These are the instructions that an object executes in order to change its state and the state of the objects it references.
\end{itemize}
%%%
Objects are abstractly defined in source code. Source code is written in a human readable/writeable language. An example \ttt{Person} class defined in the Java language is presented below. This particular class has two fields (i.e.~\ttt{age} and \ttt{friends}) and one method (i.e.~\ttt{makeFriend}).

\begin{verbatim}
  public class Person {
    int age;
    Collection<Person> friends;
    
    public void makeFriend(Person p) {
      this.friends.add(p);
    }	
  }
\end{verbatim}
%%%
There is an important distinction between a class and an object. A class is an abstract description of an object. Classes are written in source code. Object's are created during the run-time of the executed code and embody the properties of their abstract class. In this way, objects instantiate (or realize) classes. Before objects can be created, a class described in source code must be compiled so that the machine can more efficiently process the code. In other words, the underlying machine has a very specific instruction set (or language) that it uses. It is the role of the compiler to translate source code into machine-readable instructions. Instructions can be represented in the native language of the hardware processor (i.e.~according to its instruction set) or it can be represented in an intermediate language that can be processed by a virtual machine (i.e.~software that simulates the behavior of a hardware machine). If a virtual machine language is used, it is ultimately the role of the virtual machine to translate the instructions it is processing into the instruction set used by the underlying hardware machine. However, the computing stack does not end there. It is ultimately up to the ``laws of physics'' to alter the state of the hardware machine. As the hardware machine changes states, its alters the state of all the layers of abstractions built atop it.

Object-oriented programming is perhaps the most widely used software development paradigm and is part of the general knowledge of most computer scientists and engineers. Examples of the more popular object-oriented languages include C++, Java, and Python. Some of the benefits of object-oriented programming are itemized below.
%%%
\begin{itemize}\addtolength{\itemsep}{-0.5\baselineskip}
	\item abstraction: representing a problem intuitively as a set of interacting objects.
	\item encapsulation: methods and fields are ``bundled'' with particular objects.
	\item inheritance: subclasses inherit the fields and methods of their parent classes.
\end{itemize}
%%%
In general, as systems scale, the management of large bodies of code is made easier through the use of object-oriented programming.

There exist many similarities between the RDFS and OWL Semantic Web ontology languages discussed in \S \ref{sec:knowledge-base} and the typical object-oriented programming languages previously mentioned. For example, in the ontology languages, there exist the notion of classes, their instances (i.e.~objects), and instance properties (i.e.~fields).\footnote{It is noted that the semantics of inheritance and properties in object-oriented languages are different than those of RDFS and OWL. Object-oriented languages are frame-based and tend to assume a closed world \cite{frameowl:wang2006}. Also, there does not exist the notion of sub-properties in object-oriented languages as fields are not ``first-class citizens.''} However, the biggest differentiator is that objects in object-oriented environments maintain methods. The only computations that occur in RDFS and OWL are through the inference rules of the logic they implement and as such are not specific to particular classes. Even if rules are implemented for particular classes (for example, in SWRL \cite{swrl:horrocks2004}), such rule languages are not typically Turing-complete \cite{compute:turing1937} and thus, do not support general-purpose computing.

In order to bring general-purpose, object-oriented computing to the Web of Data, various object-oriented languages have been developed that represent their classes and their objects in RDF. Much like rule languages such as SWRL have an RDF encoding, these object-oriented languages do as well. However, they are general-purpose imperative languages that can be used to perform any type of computation. Moreover, they are object-oriented so that they have the benefits associated with object-oriented systems itemized previously. When human readable-writeable source code written in an RDF programming language is compiled, it is compiled into RDF. By explicitly encoding methods in RDF---their instruction-level data---the Web of Data is transformed into a Web of Process.\footnote{It is noted that the Web of Process is not specifically tied to object-oriented languages. For example, the Ripple programming language is a relational language where computing instructions are stored in \ttt{rdf:List}s \cite{ripple:shinavier2007}. Ripple is generally useful for performing complex query and insert operations on the Web of Process. Moreover, because programs are denoted by URIs, it is easy to link programs together by referencing URIs.} The remainder of this section will discuss three computing models on the Web of Process:
%%%
\begin{enumerate}\addtolength{\itemsep}{-0.5\baselineskip}
	\item partial object repository: where typical object-oriented languages utilize the Web of Process to store object field data, not class descriptions and methods.
	\item full object repository: where RDF-based object-oriented languages encode classes, object fields, and object methods in RDF.
	\item virtual machine repository: where RDF-based classes, objects, and virtual machines are represented in the Web of Process.
\end{enumerate}

\subsection{Partial Object Repository}

The Web of Process can be used as a partial object repository. In this sense, objects represented in the Web of Process only maintain their fields, not their methods. It is the purpose of some application represented external to the Web of Process to store and retrieve object data from the Web of Process. In many ways, this model is analogous to a ``black board'' tuple-space \cite{tuple:gelernter1992}.\footnote{Object-spaces such as JavaSpaces is a modern object-oriented use of a tuple-space \cite{javaspace:freeman2008}.} By converting the data that is encoded in the Web of Process to an object instance, the Web of Process serves as a database for populating the objects of an application. It is the role of this application to provide a mapping from the RDF encoded object to its object representation in the application (and vice versa for storage). A simple mapping is that a URI can denote a pointer to a particular object. The predicates of the statements that have the URI as a subject are seen as the field names. The objects of those statements are the values of those fields. For example, given the \ttt{Person} class previously defined, an instance in RDF can be represented as

\begin{verbatim}
(lanl:1234, rdf:type, lanl:Person)
(lanl:1234, lanl:age, "29"^^xsd:int)
(lanl:1234, lanl:friend, lanl:2345)
(lanl:1234, lanl:friend, lanl:3456)
(lanl:1234, lanl:friend, lanl:4567),
\end{verbatim}
%%%
where \ttt{lanl:1234} represents the \ttt{Person} object and the \ttt{lanl:friend} properties points to three different \ttt{Person} instances. This simple mapping can be useful for many types of applications. However, it is important to note that there exists a mismatch between the semantics of RDF, RDFS, and OWL and typical object-oriented languages. In order to align both languages it is possible either to 1.) ignore RDF/RDFS/OWL semantics and interpret RDF as simply a data model for representing an object or 2.) make use of complicated mechanisms to ensure that the external object-oriented environment is faithful to such semantics \cite{owljava:kalyanpur2004}.

Various RDF-to-object mappers exists. Examples include Schemagen\footnote{Schemagen is currently available at \url{http://jena.sourceforge.net/}.}, Elmo\footnote{Elmo is currently available at \url{http://www.openrdf.org/}.}, and ActiveRDF \cite{activerdf:oren2008}. RDF-to-object mappers usually provide support to 1.) automatically generate class definitions in the non-RDF language, 2.) automatically populate these objects using RDF data, and 3.) automatically write these objects to the Web of Process. With RDF-to-object mapping, what is preserved in the Web of Process is the description of the data contained in an object (i.e.~its fields), not an explicit representation of the object's process information (i.e.~its methods). However, there exists RDF object-oriented programming languages that represent methods and their underlying instructions in RDF.

\subsection{Full Object Repository}

The following object-oriented languages compile human readable/writeable source code into RDF: Adenine \cite{adenine:quan2003}, Adenosine, FABL \cite{fabl:bureau2001}, and Neno \cite{rodriguez:gpsemnet2009}. The compilation process creates a full RDF representation of the classes defined. The instantiated objects of these classes are also represented in RDF. Thus, the object fields and their methods are stored in the Web of Process. Each aforementioned RDF programming language has an accompanying virtual machine. It is the role of the respective virtual machine to query the Web of Process for objects, execute their methods, and store any changes to the objects back into the Web of Process. 

Given that these languages are designed specifically for an RDF environment and in many cases, make use of the semantics defined for RDFS and OWL, the object-oriented nature of these languages tend to be different than typical languages such as C++ and Java. Multiple inheritance, properties as classes, methods as classes, unique SPARQL-based language constructs, etc. can be found in these languages. To demonstrate methods as classes and unique SPARQL-based language constructs, two examples are provided from Adenosine and Neno, respectively. In Adenosine, methods are declared irrespective of a class and can be assigned to classes as needed.

\begin{verbatim}
(lanl:makeFriend, rdf:type, std:Method)
(lanl:makeFriend, std:onClass, lanl:Person)
(lanl:makeFriend, std:onClass, lanl:Dog).
\end{verbatim}
%%%
Next, in Neno, it is possible to make use of the inverse query capabilities of SPARQL. The Neno statement

\begin{verbatim}
rpi:josh.lanl:friend.lanl:age;
\end{verbatim}
%%%
is typical in many object-oriented languages: the age of the friends of Josh.\footnote{Actually, this is not that typical as fields cannot denote multiple objects in most object-oriented langauges. In order to reference multiple objects, fields tend to reference an abstract ``collection'' object that contains multiple objects within it (e.g.~an array).} However, the statement

\begin{verbatim}
rpi:josh..lanl:friend.lanl:age;
\end{verbatim}
%%%
is not. This statement makes use of ``dot dot'' notation and is called inverse field referencing. This particular example returns the age of all the people that are friends with Josh. That is, it determines all the \ttt{lanl:Person} objects that are a \ttt{lanl:friend} of \ttt{lanl:josh} and then returns the \ttt{xsd:int} of their \ttt{lanl:age}. This expression resolves to the SPARQL query

\begin{verbatim}
SELECT ?y WHERE {
  ?x <lanl:friend> <rpi:josh> .
  ?x <lanl:age> ?y }.
\end{verbatim}

In RDF programming languages, there does not exist the impedance mismatch that occurs when integrating typical object-oriented languages with the Web of Process. Moreover, such languages can leverage many of the standards and technologies associated with the Web of Data in general. In typical object-oriented languages, the local memory serves as the object storage environment. In RDF object-oriented languages, the Web of Process serves this purpose. An interesting consequence of this model is that because compiled classes and instantiated objects are stored in the Web of Process, RDF software can easily reference other RDF software in the Web of Process. Instead of pointers being $32$- or $64$-bit addresses in local memory, pointers are URIs. In this medium, the Web of Process is a shared memory structure by which all the world's software and data can be represented, interlinked, and executed.
%%%
\begin{quote}
``The formalization of computation within RDF allows active content to be integrated seamlessly into RDF repositories, and provides a programming environment which simplifies the manipulation of RDF when compared to use of a conventional language via an API.'' \cite{fabl:bureau2001}
\end{quote}
%%%
A collection of the previously mentioned benefits of RDF programming are itemized below.
%%%
\begin{itemize}\addtolength{\itemsep}{-0.5\baselineskip}
	\item the language and RDF are strongly aligned: there is a more direct mapping of the language constructs and the underlying RDF representation.
	\item compile time type checking: RDF APIs will not guarantee the validity of an RDF object at compile time.
	\item unique language constructs: Web of Data technology and standards are more easily adopted into RDF programming languages.
	\item reflection: language reflection is made easier because everything is represented in RDF.
	\item reuse: software can reference other software by means of URIs.
\end{itemize}

There are many issues with this model that are not discussed here. For example, issues surrounding security, data integrity, and computational resource consumption make themselves immediately apparent. Many of these issues are discussed, to varying degrees of detail, in the publications describing these languages.

\subsection{Virtual Machine Repository}

In the virtual machine repository model, the Web of Process is made to behave like a general-purpose computer. In this model, software, data, and virtual machines are all encoded in the Web of Process. The Fhat RDF virtual machine (RVM) is a virtual machine that is represented in RDF \cite{rodriguez:gpsemnet2009}. The Fhat RVM has an architecture that is similar to other high-level virtual machines such as the Java virtual machine (JVM). For example, it maintains a program counter (e.g.~a pointer to the current instruction being executed), various stacks (e.g.~operand stack, return stack, etc.), variable frames (e.g.~memory for declared variables), etc. However, while the Fhat RVM is represented in the Web of Process, it does not have the ability to alter its state without the support of some external process. An external process that has a reference to a Fhat RVM can alter it by moving its program location through a collection of instructions, by updating its stacks, by altering the objects in its heap, etc. Again, the Web of Process (and more generally, the Web of Data) is simply a data structure. While it can represent process information, it is up to machines external to the Web of Process to manipulate it and thus, alter its state.

In this computing model, a full computational stack is represented in the Web of Process. Computing, at this level, is agnostic to the physical machines that support its representation. The lowest-levels of access are URIs and their RDF relations. There is no pointer to physical memory, disks, network cards, video cards, etc. Such RDF software and RVMs exist completely in an abstract URI and RDF address space---in the Web of Process. In this way, if an external process that is executing an RVM stops, the RVM simply ``freezes'' at its current instruction location. The state of the RVM halts. Any other process with a reference to that RVM can continue to execute it.\footnote{In analogy, if the laws of physics stopped ``executing'' the world, the state of the world would ``freeze'' awaiting the process to continue.} Similarly, an RVM represented on one physical machine can compute an object represented on another physical machine. However, for the sake of efficiency, given that RDF subgraphs can be easily downloaded by a physical machine, the RVMs can be migrated between data stores---the process is moved to the data, not the data to the process. Many issues surrounding security, data integrity, and computational resource consumption are discussed in \cite{rodriguez:gpsemnet2009}. Currently there exists the concept, the consequences, and a prototype of an RVM. Future work in this area will hope to transform the Web of Process (and more generally, the Web of Data) into massive-scale, distributed, general-purpose computer.

\section{Conclusion}

A URI can denote anything. It can denote a term, a vertex, an instruction. However, by itself, a single URI is not descriptive. When a URI is interpreted within the context of other URIs and literals, it takes on a richer meaning and is more generally useful. RDF is the means of creating this context. Both the URI and RDF form the foundational standards of the Web of Data. From the perspective of the domain of knowledge representation and reasoning, the Web of Data is a distributed knowledge base---a Semantic Web. In this interpretation, according to which ever logic is used, existing knowledge can be used to infer new knowledge. From the perspective of the domain of network analysis, the Web of Data is a distributed multi-relational network---a Giant Global Graph. In this interpretation, network algorithms provide structural statistics and can support network-based information retrieval systems. From the perspective of the domain of object-oriented programming, the Web of Data is a distribute object repository---a Web of Process. In this interpretation, a complete computing environment exists that yields a general-purpose, Web-based, distributed computer. For other domains, other interpretations of the Web of Data can exist. Ultimately, the Web of Data can serve as a general-purpose medium for storing and relating all the world's data. As such, machines can usher in a new era of global-scale data management and processing.

\section*{Acknowledgements}

Joshua Shinavier of the Rensselaer Polytechnic Institute and Joe Geldart of the University of Durham both contributed through thoughtful discussions and review of the article.

\label{lastpage-01}

\begin{thebibliography}{10}

\bibitem{agraph:aasman2006}
Jans Aasman.
\newblock Allegro graph.
\newblock Technical Report~1, Franz Incorporated, 2006.

\bibitem{diamwww:albert1999}
Reka Albert and Albert-Laszlo Barabasi.
\newblock Diameter of the world wide web.
\newblock {\em Nature}, 401:130--131, September 1999.

\bibitem{oracle:alexander2006}
Nicole Alexander and Siva Ravada.
\newblock {RDF} object type and reification in the database.
\newblock In {\em {Proceedings of the International Conference on Data
  Engineering}}, pages 93--103, Washington, DC, 2006. {IEEE}.

\bibitem{spread:anderson1983}
John~R. Anderson.
\newblock A spreading activation theory of memory.
\newblock {\em Journal of Verbal Learning and Verbal Behaviour}, 22:261--295,
  1983.

\bibitem{close:bavelas1950}
Alex Bavelas.
\newblock Communication patterns in task oriented groups.
\newblock {\em The Journal of the Acoustical Society of America}, 22:271--282,
  1950.

\bibitem{lee94}
Tim Berners-Lee, Robert Cailliau, Ari Luotonen, Henrik~F. Nielsen, and Arthur
  Secret.
\newblock {The World-Wide Web}.
\newblock {\em Communications of the ACM}, 37:76--82, 1994.

\bibitem{uri:berners2005}
Tim Berners-Lee, Roy~T. Fielding, Day Software, Larry Masinter, and Adobe
  Systems.
\newblock {U}niform {R}esource {I}dentifier ({URI}): {G}eneric {S}yntax,
  January 2005.

\bibitem{lee:semantic2001}
Tim Berners-Lee, James~A. Hendler, and Ora Lassila.
\newblock The {S}emantic {W}eb.
\newblock {\em Scientific American}, pages 34--43, May 2001.

\bibitem{linkeddata:bizer2008}
Christian Bizer, Tom Heath, Kingsley Idehen, and Tim Berners-Lee.
\newblock Linked data on the web.
\newblock In {\em {P}roceedings of the {I}nternational {W}orld {W}ide {W}eb
  {C}onference}, Linked Data Workshop, Beijing, China, April 2008.

\bibitem{bollen:mesur2008}
Johan Bollen, Herbert {Van de Sompel}, and Marko~A. Rodriguez.
\newblock Towards usage-based impact metrics: first results from the {MESUR}
  project.
\newblock In {\em {Proceedings of the Joint Conference on Digital Libraries}},
  pages 231--240, New York, NY, 2008. {IEEE/ACM}.

\bibitem{diamscale:2004}
B\'{e}laa Bollob\'{a}s and Oliver Riordan.
\newblock The diameter of a scale-free random graph.
\newblock {\em Combinatorica}, 24(1):5--34, 2004.

\bibitem{netanal:brandes2005}
Ulrick Brandes and Thomas Erlebach, editors.
\newblock {\em Network Analysis: Methodolgical Foundations}.
\newblock Springer, Berling, DE, 2005.

\bibitem{betweeness:brandes2001}
Ulrik Brandes.
\newblock A faster algorithm for betweeness centrality.
\newblock {\em Journal of Mathematical Sociology}, 25(2):163--177, 2001.

\bibitem{rdfs:brickley2004}
Dan Brickley and Ramanathan~V. Guha.
\newblock {RDF} vocabulary description language 1.0: {RDF} schema.
\newblock Technical report, World Wide Web Consortium, 2004.

\bibitem{anatom:brin1998}
Sergey Brin and Lawrence Page.
\newblock The anatomy of a large-scale hypertextual web search engine.
\newblock {\em Computer Networks and ISDN Systems}, 30(1--7):107--117, 1998.

\bibitem{sesame:dutchy2002}
Jeen Broekstra, Arjohn Kampman, and Frank van Harmelen.
\newblock Sesame: A generic architecture for storing and querying {RDF}.
\newblock In {\em {P}roceedings of the {I}nternational {S}emantic {W}eb
  {C}onference}, Sardinia, Italy, June 2002.

\bibitem{graph:chartrand1977}
Gary Chartrand.
\newblock {\em Introductory Graph Theory}.
\newblock Dover, 1977.

\bibitem{spread:collins1975}
Allan~M. Collins and Elizabeth~F. Loftus.
\newblock A spreading activation theory of semantic processing.
\newblock {\em Psychological Review}, 82:407--428, 1975.

\bibitem{logic:copi1982}
Irving~M. Copi.
\newblock {\em Introduction to Logic}.
\newblock Macmillan Publishing Company, New York, NY, 1982.

\bibitem{short:dijkstra1959}
Edsger~W. Dijkstra.
\newblock A note on two problems in connexion with graphs.
\newblock {\em Numerische Mathematik}, 1:269--271, 1959.

\bibitem{larkc:fensel2008}
Dieter Fensel, Frank van Harmelen, Bo~Andersson, Paul Brennan, Hamish
  Cunningham, Emanuele~Della Valle, Florian Fischer, Zhisheng Huang, Atanas
  Kiryakov, Tony~Kyung il~Lee, Lael School, Volker Tresp, Stefan Wesner,
  Michael Witbrock, and Ning Zhong.
\newblock Towards larkc: a platform for web-scale reasoning.
\newblock In {\em {Proceedings of the IEEE International Conference on Semantic
  Computing}}, Los Alamitos, CA, 2008. {IEEE}.

\bibitem{javaspace:freeman2008}
Eric Freeman, Susanne Hupfer, and Ken Arnold.
\newblock {\em JavaSpaces: Principles, Patterns, and Practice}.
\newblock Prentice Hall, 2008.

\bibitem{between:freeman1977}
Linton~C. Freeman.
\newblock A set of measures of centrality based on betweenness.
\newblock {\em Sociometry}, 40(35--41), 1977.

\bibitem{tuple:gelernter1992}
David Gelernter and Nicholas Carriero.
\newblock Coordination languages and their significance.
\newblock {\em Communications of the {ACM}}, 35(2):97--107, 1992.

\bibitem{fabl:bureau2001}
Chris Goad.
\newblock Describing computation within {RDF}.
\newblock In {\em {Proceedings of the International Semantic Web Working
  Symposium}}, 2004.

\bibitem{racer:haarslev2003}
Volker Haarslev and Ralf M\"{o}ller.
\newblock Racer: A core inference engine for the {S}emantic {W}eb.
\newblock In {\em {Proceedings of the 2nd International Workshop on Evaluation
  of Ontology-based Tools}}, pages 27--36, 2003.

\bibitem{markov:haggstrom2002}
Olle H{\"a}ggstr{\"o}m.
\newblock {\em Finite {M}arkov Chains and Algorithmic Applications}.
\newblock Cambridge University Press, 2002.

\bibitem{eccen:harary1995}
Frank Harary and Per Hage.
\newblock Eccentricity and centrality in networks.
\newblock {\em Social Networks}, 17:57--63, 1995.

\bibitem{rdfsem:hayes2004}
Patrick Hayes and Brian McBride.
\newblock {RDF} semantics.
\newblock Technical report, World Wide Web Consortium, February 2004.

\bibitem{neural:haykin1999}
Simon Haykin.
\newblock {\em Neural Networks. A Comprehensive Foundation}.
\newblock Prentice Hall, New Jersey, USA, 1999.

\bibitem{matrix:horn1994}
Roger Horn and Charles Johnson.
\newblock {\em Topics in Matrix Analysis}.
\newblock Cambridge University Press, 1994.

\bibitem{swrl:horrocks2004}
Ian Horrocks, Peter~F. Patel-Schneider, Harold Boley, Said Tabet, Benjamin
  Grosof, and Mike Dean.
\newblock {SWRL}: A {S}emantic {W}eb rule language combining {OWL} and
  {RuleML}.
\newblock Technical report, World Wide Web Consortium, May 2004.

\bibitem{owljava:kalyanpur2004}
A.~Kalyanpur, D.~Pastor, S.~Battle, and J.~Padget.
\newblock Automatic mapping of {OWL} ontologies into java.
\newblock In {\em {Proceedings of Software Engineering. - Knowledge
  Engineering}}, Banff, Canada, 2004.

\bibitem{owlim:kiryakov2005}
Atanas Kiryakov, Damyan Ognyanov, and Dimitar Manov.
\newblock {OWLIM} -- a pragmatic semantic repository for {OWL}.
\newblock In {\em {I}nternational {W}orkshop on {S}calable {S}emantic {W}eb
  {K}nowledge {B}ase {S}ystems}, volume LNCS 3807, pages 182--192, New York,
  NY, November 2005. Spring-Verlag.

\bibitem{advance:kosch2004}
Dirk Kosch\"{u}tzki, Katharina~Anna Lehmann, Dagmar Tenfelde-Podehl, and Oliver
  Zlotowski.
\newblock {\em Network Analysis: Methodolgical Foundations}, volume 3418 of
  {\em Lecture Notes in Computer Science}, chapter Advanced Centrality
  Concepts, pages 83--111.
\newblock Spring-Verlag, 2004.

\bibitem{owl:lacy2005}
Lee~W. Lacy.
\newblock {\em {OWL}: Representing Information Using the {W}eb {O}ntology
  {L}anguage}.
\newblock Trafford Publishing, 2005.

\bibitem{uuid:leach2005}
Paul~J. Leach.
\newblock A {U}niversally {U}nique {ID}entifier ({UUID}) {URN} {N}amespace.
\newblock Technical report, Network Working Group, 2005.

\bibitem{close:leavitt1951}
Harold~J. Leavitt.
\newblock Some effects of communication patterns on group performance.
\newblock {\em Journal of Abnornal and Social Psychology}, 46:38--50, 1951.

\bibitem{owlspec:mcguinness2004}
Deborah~L. McGuinness and Frank van Harmelen.
\newblock {OWL} web ontology language overview, February 2004.

\bibitem{newman:assort}
Mark E.~J. Newman.
\newblock Assortative mixing in networks.
\newblock {\em Physical Review Letters}, 89(20):208701, 2002.

\bibitem{newman:mixpatt2003}
Mark E.~J. Newman.
\newblock Mixing patterns in networks.
\newblock {\em Physical Review E}, 67(2):026126, Feb 2003.

\bibitem{activerdf:oren2008}
Eyal Oren, Benjamin Heitmann, and Stefan Decker.
\newblock Active{RDF}: Embedding semantic web data into object-oriented
  languages.
\newblock {\em Web Semantics: Science, Services and Agents on the World Wide
  Web}, 6(3):191--202, 2008.

\bibitem{gene:ozgur2008}
Arzucan Ozgur, Thuy Vu, Gunes Erkan, and Dragomir~R. Radev.
\newblock Identifying gene-disease associations using centrality on a
  literature mined gene-interaction network.
\newblock {\em Bioinformatics}, 24(13):277--285, July 2008.

\bibitem{pellet:2004}
Bijan Parsia and Evren Sirin.
\newblock {P}ellet: An {OWL DL} reasoner.
\newblock In {\em {P}roceedings of the {I}nternational {S}emantic {W}eb
  {C}onference}, volume 3298 of {\em Lecture Notes in Computer Science},
  Hiroshima, Japan, November 2004. Springer-Verlag.

\bibitem{syllogism:patzig1968}
Gunther Patzig.
\newblock {\em Aristotle's Theory of The Syllogism}.
\newblock D. Reidel Publishing Company, Boston, Massachusetts, 1968.

\bibitem{sparql:prud2004}
Eric Prud'hommeaux and Andy Seaborne.
\newblock {SPARQL} query language for {RDF}.
\newblock Technical report, World Wide Web Consortium, October 2004.

\bibitem{adenine:quan2003}
Dennis Quan, David~F. Huynh, Vineet Sinha, and David Karger.
\newblock Adenine: A metadata programming language.
\newblock Technical report, Massachusetts Institute of Technology, February
  2003.

\bibitem{grammar:rodriguez2008}
Marko~A. Rodriguez.
\newblock Grammar-based random walkers in semantic networks.
\newblock {\em Knowledge-Based Systems}, 21(7):727--739, 2008.

\bibitem{rodriguez:gpsemnet2009}
Marko~A. Rodriguez.
\newblock {\em Emergent Web Intelligence}, chapter General-Purpose Computing on
  a Semantic Network Substrate.
\newblock Springer-Verlag, Berlin, DE, 2009.

\bibitem{pathalg:rodriguez2008}
Marko~A. Rodriguez and Joshua Shinavier.
\newblock Exposing multi-relational networks to single-relational network
  analysis algorithms.
\newblock Technical Report LA-UR-08-03931, Los Alamos National Laboratory,
  2008.

\bibitem{geodesics:rodriguez2007}
Marko~A. Rodriguez and Jennifer~H. Watkins.
\newblock Grammar-based geodesics in semantic networks.
\newblock Technical Report LA-UR-07-4042, Los Alamos National Laboratory, 2007.

\bibitem{close:sabaidussi1966}
Gert Sabidussi.
\newblock The centrality index of a graph.
\newblock {\em Psychometrika}, 31:581--603, 1966.

\bibitem{ripple:shinavier2007}
Joshua Shinavier.
\newblock Functional programs as linked data.
\newblock In {\em 3rd {W}orkshop on {S}cripting for the {S}emantic {W}eb},
  Innsbruck, Austria, 2007.

\bibitem{compute:turing1937}
Alan~M. Turing.
\newblock On computable numbers, with an application to the
  entscheidungsproblem.
\newblock {\em Proceedings of the London Mathematical Society}, 42(2):230--265,
  1937.

\bibitem{uri:2001}
{W3C/IETF}.
\newblock {URI}s, {URL}s, and {URN}s: Clarifications and recommendations 1.0,
  September 2001.

\bibitem{frameowl:wang2006}
Hai~H. Wang, Natasha Noy, Alan Rector, Mark Musen, Timothy Redmond, Daniel
  Rubin, Samson Tu, Tania Tudorache, Nick Drummond, Matthew Horridge, and
  Julian Sedenberg.
\newblock Frames and {OWL} side by side.
\newblock In {\em 10th {I}nternational {P}rot{\'e}g{\'e} {C}onference},
  Budapest, Hungary, July 2007.

\bibitem{nars2.2:wang}
Pei Wang.
\newblock Non-axiomatic reasoning system (version 2.2).
\newblock Technical Report~75, Center for Research on Concepts and Cognition at
  Indiana University, 1993.

\bibitem{inherit:wang1994}
Pei Wang.
\newblock From inheritance relation to non-axiomatic logic.
\newblock {\em International Journal of Approximate Reasoning}, 11:281--319,
  1994.

\bibitem{coglogic:wang2004}
Pei Wang.
\newblock Cognitive logic versus mathematical logic.
\newblock In {\em Proceedings of the {Third International Seminar on Logic and
  Cognition}}, May 2004.

\bibitem{nal:wang2006}
Pei Wang.
\newblock {\em Rigid Flexibility: The Logic Of Intelligence}.
\newblock Springer, 2006.

\bibitem{socialanal:wasserman1994}
Stanley Wasserman and Katherine Faust.
\newblock {\em Social Network Analysis: Methods and Applications}.
\newblock Cambridge University Press, Cambridge, UK, 1994.

\bibitem{markov:white2003}
Scott White and Padhraic Smyth.
\newblock Algorithms for estimating relative importance in networks.
\newblock In {\em {Proceedings of the International Conference on Knowledge
  Discovery and Data Mining}}, pages 266--275, New York, NY, 2003. ACM Press.

\end{thebibliography}
\end{document}